\documentclass[10pt,twocolumn,letterpaper]{article}
\usepackage[pagenumbers]{cvpr}

\usepackage{graphicx}
\usepackage[fleqn]{amsmath}
\usepackage{amssymb}
\usepackage{booktabs}
\usepackage{url}            
\usepackage{booktabs}       
\usepackage{amsfonts}       
\usepackage{nicefrac}       
\usepackage{microtype}      
\usepackage{xcolor}         
\usepackage {enumitem}
\usepackage{multirow}
\setlist {nolistsep}
\usepackage{makecell}
\usepackage{comment}
\usepackage[pagebackref,breaklinks,colorlinks]{hyperref}
\hypersetup{colorlinks=true}

\usepackage[capitalize]{cleveref}
\crefname{section}{Sec.}{Secs.}
\Crefname{section}{Section}{Sections}
\Crefname{table}{Table}{Tables}
\crefname{table}{Tab.}{Tabs.}

\def\cvprPaperID{xxxx}
\def\confName{CVPR}
\def\confYear{2023}

\begin{document}

\title{Referring Image Matting}

\author{Jizhizi Li, Jing Zhang, and Dacheng Tao\\
The University of Sydney, Sydney, Australia\\
{\tt\small jili8515@uni.sydney.edu.au, jing.zhang1@sydney.edu.au, dacheng.tao@gmail.com}
\thanks{Dr Jing Zhang and Ms Jizhizi Li were supported by Australian Research Council Projects in part by FL170100117 and IH180100002.}}

\maketitle

\begin{abstract}
Different from conventional image matting, which either requires user-defined scribbles/trimap to extract a specific foreground object or directly extracts all the foreground objects in the image indiscriminately, we introduce a new task named \textbf{Referring Image Matting (RIM)} in this paper, which aims to extract the meticulous alpha matte of the specific object that best matches the given natural language description, thus enabling a more natural and simpler instruction for image matting. First, we establish a large-scale challenging dataset \textbf{RefMatte} by designing a comprehensive image composition and expression generation engine to automatically produce high-quality images along with diverse text attributes based on public datasets. RefMatte consists of 230 object categories, 47,500 images, 118,749 expression-region entities, and 474,996 expressions. Additionally, we construct a real-world test set with 100 high-resolution natural images and manually annotate complex phrases to evaluate the out-of-domain generalization abilities of RIM methods. Furthermore, we present a novel baseline method \textbf{CLIPMat} for RIM, including a context-embedded prompt, a text-driven semantic pop-up, and a multi-level details extractor. Extensive experiments on RefMatte in both keyword and expression settings validate the superiority of CLIPMat over representative methods. We hope this work could provide novel insights into image matting and encourage more follow-up studies. The dataset, code and models are available at \href{https://github.com/JizhiziLi/RIM}{https://github.com/JizhiziLi/RIM}.
\end{abstract}

\section{Introduction}
Image matting refers to extracting the soft alpha matte of the foreground in natural images, which is beneficial for various downstream applications such as video conferences, advertisement production, and e-Commerce promotion~\cite{zhang2020empowering}. Typical matting methods can be divided into two groups: 1) the methods based on auxiliary inputs, \eg, scribble~\cite{levin2007closed} and trimap~\cite{levin2007closed,cai2019disentangled}, and 2) automatic matting methods that can extract the foreground without any human intervention~\cite{gfm,dapm}. However, the former are not applicable for fully automatic scenarios, while the latter are limited to specific categories, \eg, human~\cite{shm,mgmatting,p3mj}, animal~\cite{gfm}, or the salient objects~\cite{lf,hatt}. It is still unexplored to carry out controllable image matting on arbitrary objects based on language instructions, \eg, extracting the alpha matte of the specific object that best matches the given language description.

Recently, language-driven tasks such as referring expression segmentation (RES)~\cite{mattnet}, referring image segmentation (RIS)~\cite{Huang2020ReferringIS,Ye2019CrossModalSN,Liu2017RecurrentMI}, visual question answering (VQA)~\cite{gan2017vqs}, and referring expression comprehension (REC)~\cite{luo2020multi} have been widely studied. Great progress in these areas has been made based on many datasets like ReferIt~\cite{referit}, Google RefExp~\cite{googleref}, RefCOCO~ \cite{refcoco}, VGPhraseCut~\cite{phrasecut}, and Cops-Ref~\cite{Chen2020CopsRefAN}. However, due to the limited resolution of available datasets, visual grounding methods are restricted to the coarse segmentation level. Besides, most of the methods~\cite{clipseg,mdetr} neglect pixel-level text-visual alignment and cannot preserve sufficient details, making them difficult to be used in scenarios that require meticulous alpha mattes.

\begin{figure*}[!tbp]
  \centering
  \begin{minipage}[b]{0.48\textwidth}
    \includegraphics[width=\textwidth]{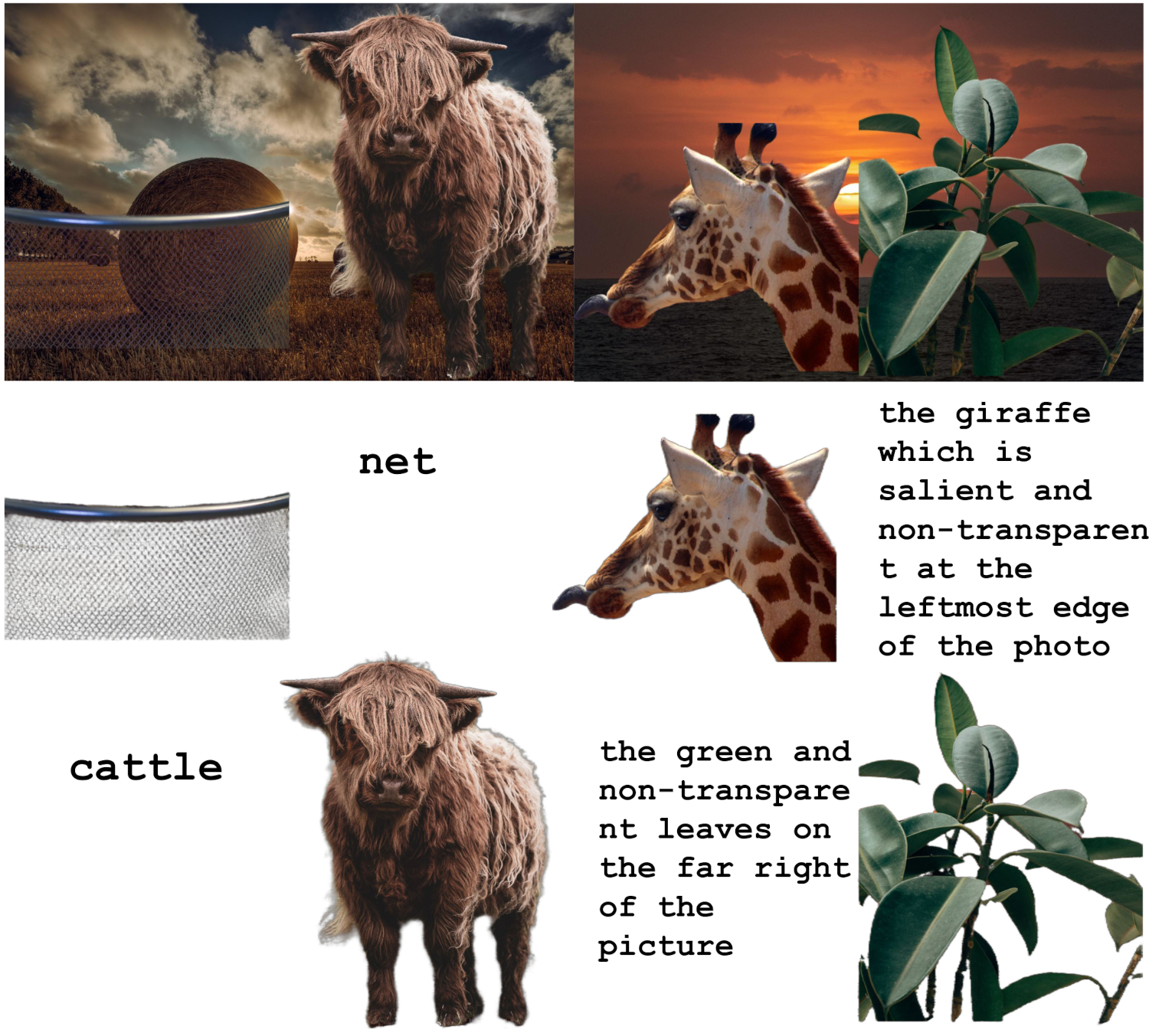}
    \caption{Some examples from our RefMatte test set (top) and the results of CLIPMat given keyword and expression inputs (bottom).}
    \label{fig:refmatte-test}
  \end{minipage}
  \hfill
  \hspace{4pt}
  \begin{minipage}[b]{0.48\textwidth}
    \includegraphics[width=\textwidth]{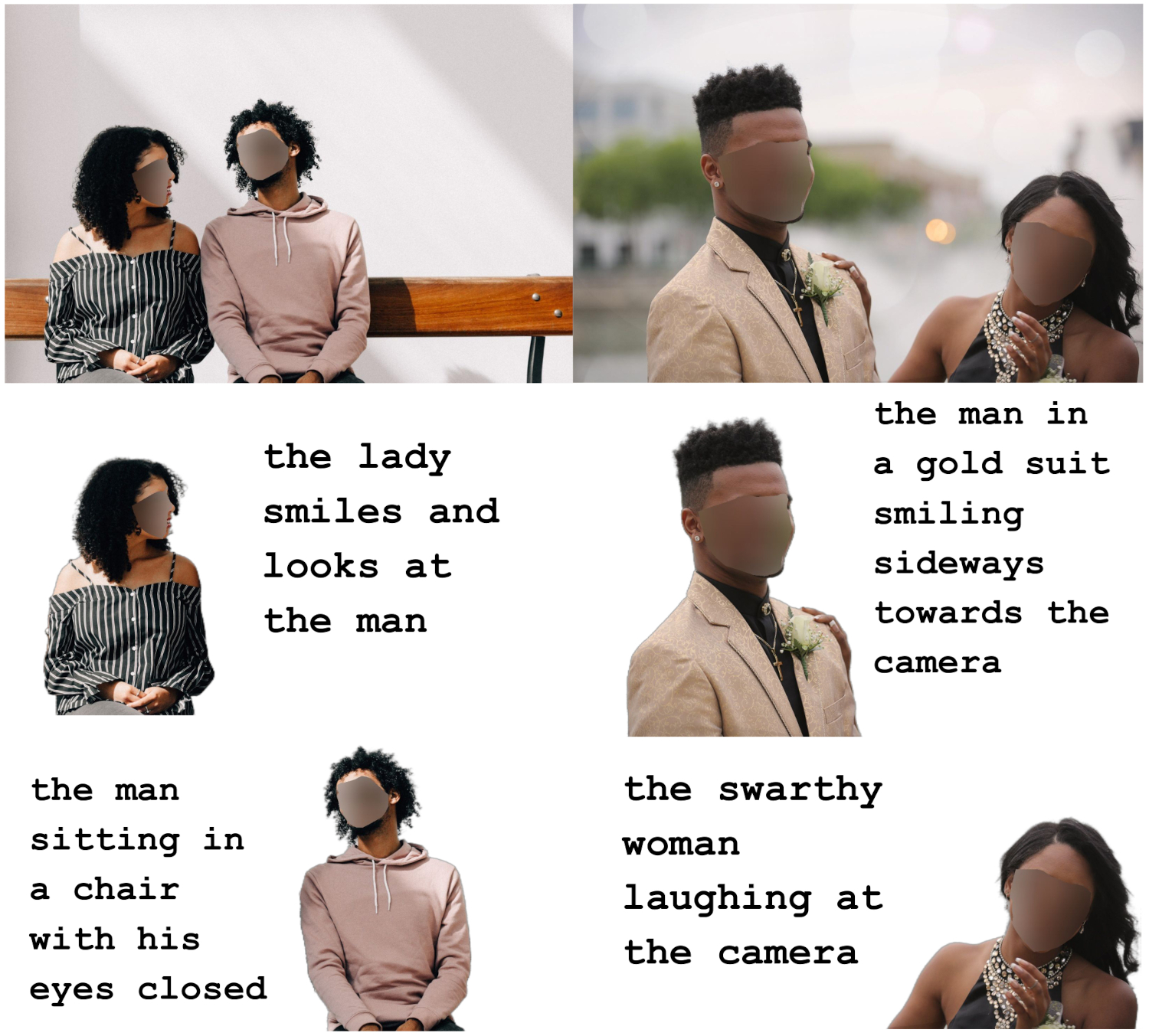}
    \caption{Some examples from our RefMatte-RW100 test set (top) and the results of CLIPMat given expression inputs (bottom), which also show CLIPMat's robustness to preserved privacy information.}
    \label{fig:refmatte-rw100}
  \end{minipage}
\end{figure*}

To fill this gap, we propose a new task named \textbf{Referring Image Matting (RIM)}, which refers to extracting the meticulous high-quality alpha matte of the specific foreground object that can best match the given natural language description from the image. Different from the conventional matting methods, RIM is designed for controllable image matting that can perform a more natural and simpler instruction to extract arbitrary objects. It is of practical significance in industrial application domains and opens up a new research direction
To facilitate the study of RIM, we establish the first dataset \textbf{RefMatte}, which consists of 230 object categories, 47,500 images, and 118,749 expression-region entities together with the corresponding high-quality alpha mattes and 474,996 expressions. Specifically, to build up RefMatte, we revisit a lot of prevalent public matting datasets like AM-2k~\cite{gfm}, P3M-10k~\cite{p3m}, AIM-500~\cite{aim}, SIM~\cite{sim} and manually label the category of each foreground object (a.k.a. entity) carefully. We also adopt multiple off-the-shelf deep learning models~\cite{mmfashion,wu2019detectron2} to generate various attributes for each entity, e.g., gender, age, and clothes type of human. Then, we design a comprehensive composition and expression generation engine to produce the synthetic images with reasonable absolute and relative positions considering other entities. Finally, we present several expression logic forms to generate varying language descriptions with the use of rich visual attributes. In addition, we propose a real-world test set RefMatte-RW100 with 100 images containing diverse objects and human-annotated expressions, which is used to evaluate the generalization ability of RIM methods. Some examples are shown in Figure~\ref{fig:refmatte-test} and Figure~\ref{fig:refmatte-rw100}.

Since previous visual grounding methods are designed for the segmentation-level tasks, directly applying them~\cite{Rao2021DenseCLIPLD,clipseg,mdetr} to the RIM task cannot produce promising alpha mattes with fine details. Here, we present CLIPMat, a novel baseline method specifically designed for RIM. CLIPMat utilizes the large-scale pre-trained CLIP~\cite{clip} model as the text and visual backbones, and the typical matting branches~\cite{gfm, p3m} as the decoders. An intuitive context-embedded prompt is adopted to provide matting-related learnable features for the text encoder. To extract high-level visual semantic information for the semantic branch, we pop up the visual semantic feature through the guidance of the text output feature. Additionally, as RIM requires much more visual details compared to the segmentation task, we devise a module to extract multi-level details by exploiting shallow-layer features and the original input image, aiming to preserve the foreground details in the matting branch. Figure~\ref{fig:refmatte-test} and Figure~\ref{fig:refmatte-rw100} show some promising results of the proposed CLIPMat given different types of language inputs, \ie, keywords and expressions.

Furthermore, to provide a fair and comprehensive evaluation of CLIPMat and relevant state-of-the-art methods, we conduct extensive experiments on RefMatte under two different settings, \ie, the keyword-based setting and expression-based setting, depending on language descriptions' forms. Both the subjective and objective results have validated the superiority of CLIPMat over representative methods. The main contribution of this study is three-fold. 1) We define a new task named RIM, aiming to identify and extract the alpha matte of the specific foreground object that best matches the given natural language description. 2) We establish the first large-scale dataset RefMatte, consisting of 47,500 images and 118,749 expression-region entities with high-quality alpha mattes and diverse expressions. 3) We present a novel baseline method CLIPMat specifically designed for RIM, which achieves promising results in two different settings of RefMatte, also on real-world images.

\section{Related Work}

\noindent\textbf{Image matting} Image matting is a fundamental computer vision task and essential for various potential downstream applications~\cite{cho2016automatic,lu2021omnimatte,dou2022learning}. Previous matting methods are divided into two groups depending on whether or not they use auxiliary user inputs. In the first group, the methods use a three-class trimap~\cite{dim,gca}, sparse scribbles~\cite{levin2007closed}, a background image~\cite{backgroundmattingv2}, a coarse map~\cite{mgmatting}, or user click~\cite{Wei2020ImprovedIM} as the auxiliary input to guide alpha estimating. Among them, scribble and click-based methods are more controllable since they usually indicate one specific foreground. However, the flexibility of these methods is still limited since the predictions are usually performed with low-level color propagation and are very sensitive to the scribbles' density~\cite{Lin2016ScribbleSupSC,p3m}. In the second group, the methods~\cite{hatt, lf, shm, gfm, aim, modnet, p3m} automatically extract the foreground objects without any manual efforts. Recently, there is also some work making efforts to control the matting process by determining which objects can be extracted. For example, Xu \etal~\cite{xu2021virtual} propose to extract the foreground human and all related objects automatically for human-object interaction. Sun \textit{et al.} propose to extract each human instance separately rather than extracting all of them indiscriminately~\cite{sun2022human}. However, it is still unexplored for controllable image matting, especially by using natural language description as guidance to extract specific foreground object that best matches the input text, even though it is efficient and flexible for the matting model to interact with a human. In this paper, we fill this gap by proposing the RIM task, the RefMatte dataset, and the baseline method CLIPMat.

\noindent\textbf{Matting datasets} Many matting datasets have been proposed to advance the progress in the image matting area. Typical matting datasets contain high-resolution images belonging to some specific object categories that have lots of details like hair, accessories, fur, and net, as well as transparent objects. For example, the matting datasets proposed by Xu \etal~\cite{dim}, Qiao \etal~\cite{hatt}, Sun \etal~\cite{sim}, and Li \etal~\cite{aim}, contain many different categories of objects, including human, animals, cars, plastic bags, and plants. Besides, some other matting datasets focus on a specific category of object, \eg, humans in P3M-10K~\cite{p3m} and animals in AM-2K~\cite{gfm}. In addition to the foreground objects, background images are also helpful for generating abundant composite images. For example, Li \etal~\cite{gfm} propose a large-scale background dataset containing 20k high-resolution and diverse images, which are helpful to reduce the domain gap between composites and natural ones. All the above datasets have open licenses and can serve as valuable resources to construct customized matting datasets, \eg, the proposed RefMatte.

Besides, it is noteworthy that due to the laborious and costly labeling process of matting datasets, existing public matting datasets~\cite{hatt,lf,sim} usually provide only the extracted foregrounds through chroma keying~\cite{dim} without the original backgrounds. To compose a reasonable amount of trainable data, a typical solution in previous matting methods~\cite{modnet,shmc,mgmatting} is to generate synthetic images like in other tasks~\cite{dwibedi2017cut,mao2022towards} by pasting the foregrounds with numerous background images. As for the domain gap between the real-world images and the composite ones, some works~\cite{gfm,modnet} have already reduced it to an acceptable range through some augmentation strategies. Although some work also present real-world matting datasets, they all contain only one foreground from a specific type, \eg, person~\cite{p3m}, animal~\cite{gfm}, or objects~\cite{hatt}, making them unsuitable to serve as the benchmark for RIM. In our work, we follow the composition route in generating RefMatte and ensure its large scale, diversity, difficulty, and high quality by synthesizing a large number of images, where there are multiple foreground objects with similar semantics and fine details on diverse backgrounds. Furthermore, we present a real-world test set with flowery human annotated expression labels to validate models' out-of-domain generalization abilities.

\noindent\textbf{Vision-language tasks and methods} Vision-Language tasks, such as RIS~\cite{Huang2020ReferringIS}, RES~\cite{mattnet}, REC~\cite{luo2020multi}, text-driven manipulation~\cite{Patashnik2021StyleCLIPTM,zhang2022promptpose}, and text-to-image generation~\cite{qiao2019mirrorgan,Qiao2019LearnIA,Ramesh2021ZeroShotTG}, have been widely studied, which are helpful for many applications like interactive image editing. Among them, RIS aims to segment the target object given language expression, which is most related but totally different from our work. The relevant methods can be divided into single-stage~\cite{Liu2017RecurrentMI,Li2018ReferringIS,clipseg,cris,Rao2021DenseCLIPLD} and two-stage ones~\cite{mdetr,mattnet,Hu2017ModelingRI,Liu2019ImprovingRE}. The former directly train a segmentation network on top of the pre-trained models like CLIP~\cite{clip}, and the latter perform sequential region proposal and segmentation. However, due to the task setting (\ie, for segmentation rather than matting) and the lack of high-quality annotations (\eg, alpha mattes)~\cite{referit,googleref,refcoco,phrasecut}, most of them have neglected the pixel-level text-semantic alignment and cannot produce fine-grained mask. Thus, we propose the new task RIM with the dataset RefMatte to facilitate the research of natural language guided image matting. Moreover, the proposed method CLIPMat with specifically designed modules could produce high-quality alpha matte and thus serve as the baseline for RIM.

\section{The RefMatte Dataset}

\begin{figure*}[t]
    \centering
    \includegraphics[width=\linewidth]{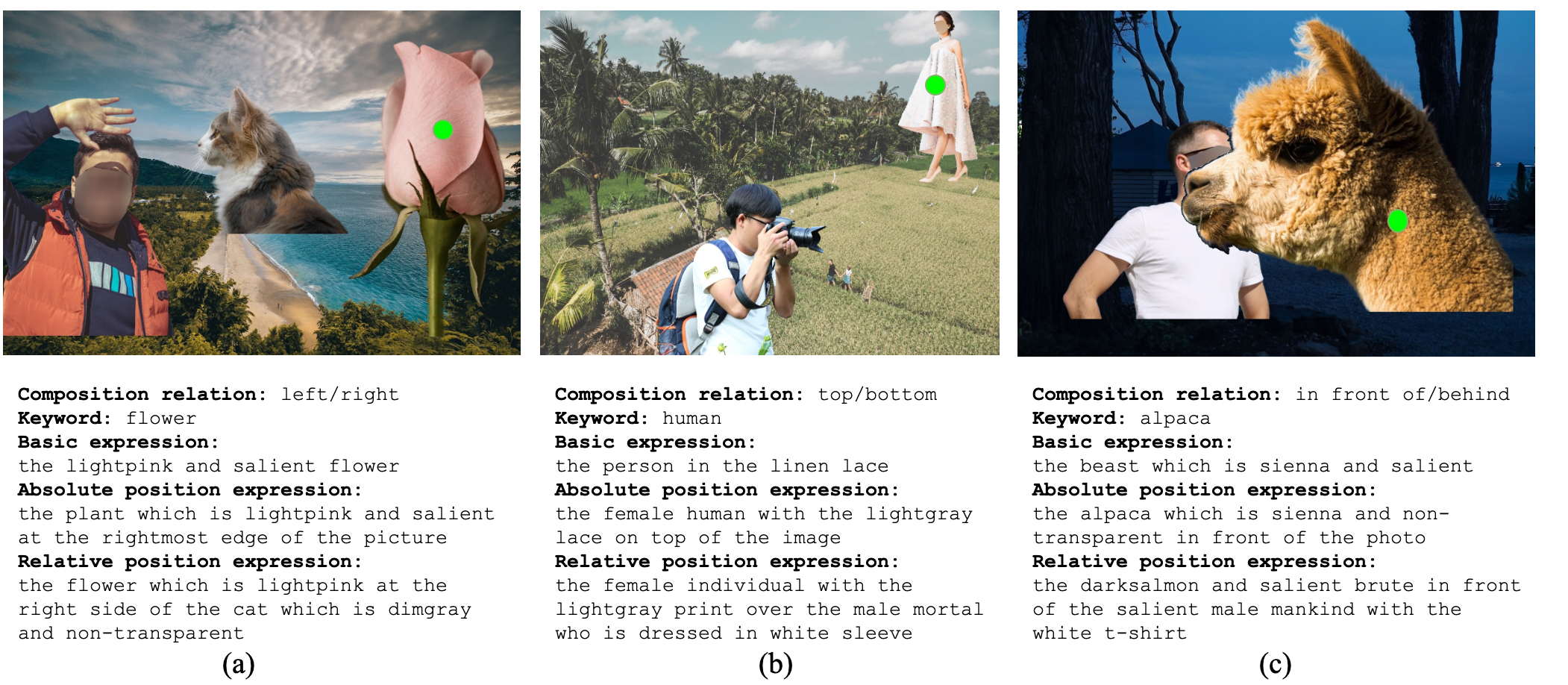}
    \caption{Some examples from our RefMatte dataset. The first row shows the composite images with different foreground instances while the second row shows the natural language descriptions corresponding to the specific foreground instances indicated by the green dots.}
    \label{fig:refmatte}
\end{figure*}

In this section, we present the overview pipeline of constructing RefMatte (Sec.~\ref{subsec:entities} and Sec.~\ref{subsec:comp}), the task settings, and a real-world test set (Sec.~\ref{subsec:tasksettings}). Figure~\ref{fig:refmatte} shows some examples from RefMatte.

\subsection{Preparation of Matting Entities}
\label{subsec:entities}

To prepare high-quality matting entities for constructing RefMatte, we revisit available matting datasets to select the required foregrounds. We then manually label each entity's category and annotate the attributes by leveraging off-the-shelf deep learning models~\cite{mmfashion,wu2019detectron2}. We present key details as follows, while more in the supplementary materials.

\noindent\textbf{Pre-processing and filtering} Due to the nature of the image matting task, all the candidate entities should be in high resolution, with clear and fine details in the alpha matte. Moreover, the data should be publicly available with open licenses and without privacy concerns. With regard to these requirements, we adopt all the foreground images from AM-2k~\cite{gfm}, P3M-10k~\cite{p3m}, and AIM-500~\cite{aim}. For other available datasets like SIM~\cite{sim}, DIM~\cite{dim}, and HATT~\cite{hatt}, we filter out those foreground images with identifiable faces in human instances and those in low-resolution or having low-quality alpha mattes. The final number of foreground entities is 13,187 in total, and we use images from BG-20k~\cite{gfm} as the background images for composition.

\noindent\textbf{Annotate the category names of entities} Previous matting datasets do not provide the specific (category) name for each entity since those matting methods extract all the objects indiscriminately. However, we need the entity name in the RIM task to describe the foreground. Following~\cite{Ordonez2013FromLS}, we label the \textit{entry-level category name} for each entity, which stands for the most commonly used name by people. Here, we adopt a semi-automatic strategy. Specifically, we use the pre-trained Mask RCNN detector~\cite{he2017mask} with a ResNet-50-FPN~\cite{he2016deep} backbone from~\cite{wu2019detectron2} to automatically detect and label the category names for each foreground instance and then manually check and correct them. In total, we have 230 categories in RefMatte. Furthermore, we adopt WordNet~\cite{Miller1992WordNetAL} to generate synonyms for each category name to enhance the diversity. We manually check the synonyms and replace some of them with more reasonable ones.

\noindent\textbf{Annotate the attributes of entities} To ensure all the entities have rich visual properties to support forming abundant expressions, we annotate them with several attributes, \eg, color for all entities, gender, age, and clothes type for the human entities. A semi-automatic strategy is adopted in retrieving such attributes. For attribute color, we cluster all the pixel values of the foreground image, find the most frequent value, and match it with the specific color in webcolors. For gender and age, we adopt the pre-trained models provided by Levi \etal in~\cite{Levi2015AgeAG} and follow common sense to define the age group based on the predicted ages. For clothes type, we adopt the off-the-shelf model provided by Liu \etal in~\cite{mmfashion}. Furthermore, motivated by the categorization of matting foregrounds in~\cite{aim}, we add the attributes of whether or not salient or transparent for all the entities as they also matter in image matting. In summary, we have at least three attributes for each entity and six attributes for human entities.

\subsection{Image Composition and Expression Generation}
\label{subsec:comp}
Based on these collected entities, we propose an image composition engine and an expression generation engine to construct RefMatte. In order to present reasonably looking composite images with semantically clear, grammatically correct, as well as abundant and fancy expressions, how to arrange the candidate entities and build up the language descriptions is the key to constructing RefMatte, which is also challenging. To this end, we define six types of position relationships for arranging entities in a composite image and leverage diverse logic forms to produce appropriate expressions. We present the details as follows. 

\noindent\textbf{Image composition engine} We adopt two or three entities for each composite to keep the entities at high resolution while arranging them with a reasonable position relationship. We define six kinds of position relationships: \textit{left, right, top, bottom, in front of,} and \textit{behind}. For each relationship, we generate the foregrounds by~\cite{levin2007closed} and composite them with the backgrounds from BG-20k~\cite{gfm} via alpha blending. Specifically, for the relationships \textit{left, right, top}, and \textit{bottom}, we ensure there are no occlusions in the instances to preserve their details. For the relationships \textit{in front of} and \textit{behind}, we simulate occlusions between the foreground instances by adjusting their relative positions. We prepare a bag of candidate words to denote each relationship and present in the supplementary materials. Some examples are in Figure~\ref{fig:refmatte}.

\noindent\textbf{Expression generation engine} To provide abundant expressions for the entities in the composite images, we define three types of expressions for each entity regarding different logic forms, where \texttt{<$att_i$>} is the attribute, \texttt{<$obj_0$>} is the category name, and \texttt{<$rel_i$>} is the relationship between the reference entity and the related one \texttt{<$obj_i$>}:

\begin{enumerate}[leftmargin=0.5cm]
  \item \textbf{\textit{Basic expression}} This is the expression that describes the target entity with as many attributes as one can, e.g, \texttt{the/a <$att_0$> <$att_1$>...<$obj_0$>} or \texttt{the/a <$obj_0$> which/that is <$att_0$> <$att_1$>, and <$att_2$>}. For example, as shown in Figure~\ref{fig:refmatte}(a), the basic expression for the entity flower is `\texttt{the lightpink and salient flower}';
  \item \textbf{\textit{Absolute position expression}} This is the expression that describes the target entity with many attributes and its absolute position in the image, \eg, \texttt{the/a <$att_0$> <$att_1$>...<$obj_0$> <$rel_0$> the photo/image/picture} or \texttt{the/a <$obj_0$> which/that is <$att_0$> <$att_1$> <$rel_0$> the photo/image/picture}. For example, as shown in Figure~\ref{fig:refmatte}(a), the absolute position expression for the flower is `\texttt{the plant which is lightpink and salient at the rightmost edge of the picture}';
  \item \textbf{\textit{Relative position expression}} This is the expression that describes the target entity with many attributes and its relative position with another entity, \eg, \texttt{the/a <$att_0$> <$att_1$>...<$obj_0$> <$rel_0$> the/a <$att_2$> <$att_3$>...<$obj_1$>} or \texttt{the/a <$obj_0$> which/that is <$att_0$> <$att_1$> <$rel_0$> the/a <$obj_1$> which/that is <$att_2$> <$att_3$>}. For example, as shown in Figure~\ref{fig:refmatte}(a), the relative position expression for the flower is `\texttt{the flower which is lightpink at the right side of the cat which is dimgray and non-transparent}'.
\end{enumerate}

\subsection{Dataset Split and Task Settings}
\label{subsec:tasksettings}

In total, We have 13,187 matting entities. We split out 11,799 for constructing the training set and 1,388 for the test set. For the training/test split, we reserve the original split in the source matting datasets except for moving all the long-tailed categories to the training set. However, the categories are not balanced since most of the entities belong to the human or animal categories. The proportion of humans, animals, and objects is 9186:1800:813 in the training set and 977:200:211 in the test set. To balance the categories, we duplicate some entities to modify the proportion to 5:1:1, leading to 10550:2110:2110 in the training set and 1055:211:211 in the test set. We then pick 5 humans, 1 animal, and 1 object as one group and feed them into the composition engine to generate an image in RefMatte. For each group in the train split, we composite 20 images with various backgrounds. For the one in the test split, we composite 10 images. The ratio of relationships \textit{left/right}:\textit{top/bottom}:\textit{in front of/behind} is set to 7:2:1. The number of entities in each image is set to 2 or 3 but fixed to 2 for relationships \textit{front of/behind} to preserve each entities' high resolution. Finally, we have 42,200 training and 2,110 test images. To further enhance the diversity of the composite images, we randomly choose entities and relationships from all candidates to form another 2,800 training images and 390 test images. Finally, we have 45,000 training images and 2,500 test images.

\noindent\textbf{Task settings} To benchmark RIM methods given different forms of language descriptions, we set up two settings upon RefMatte. We present their details as follows:
\begin{enumerate}[leftmargin=0.5cm]
  \item \textbf{\textit{keyword-based setting}} The text description in this setting is the keyword, which is the entry-level category name of the entity, \eg, \textit{flower}, \textit{human}, and \textit{alpaca} in Figure~\ref{fig:refmatte}. Please note that we filter out images with ambiguous semantic entities for this setting;
  \item \textbf{\textit{Expression-based setting}} The text description in this setting is the generated expression chosen from the basic expressions, absolute position expressions, and relative position expressions, as seen in Figure~\ref{fig:refmatte}.
\end{enumerate}

\begin{table}[htbp]
\begin{center}
\caption{Statistics of RefMatte and RefMatte-RW100.}
\label{tab:statistics}
\resizebox{\linewidth}{!}{
\begin{tabular}{c|c|ccc|c|cc|c}
\hline
Dataset  &Split & \makecell[c]{Image \\Num.} &\makecell[c]{Matte \\Num.}& \makecell[c]{Text \\Num.} &\makecell[c]{Category \\Num.} & \makecell[c]{Text \\Length} \\
\hline
  RefMatte & train & 30,391 & 77,849 & 77,849 & 230 & 1.06\\
  Keyword&  test & 1,602 & 4,085 & 4,085 & 66 & 1.04\\
  \hline
  RefMatte & train & 45,000 & 112,506 & 449,624 & 230  & 16.86\\
  Expression & test & 2,500 & 6,243 & 24,972 & 66 & 16.80 \\
 \cline{1-7}
 RefMatte-RW100 & test & 100 & 221 & 884 & 29 & 12.01\\
 \hline
\end{tabular}}
\end{center}
\end{table}

\noindent\textbf{Real-world test set} Since RefMatte is built upon composite images, a domain gap may exit when applying the models to real-world images. To further investigate the out-of-domain generalization ability of RIM models, we establish a real-world test set \textbf{RefMatte-RW100}, which consists of 100 high-resolution natural images with 2 to 3 entities in each image. The expressions are annotated by specialists following the same rules in Sec.~\ref{subsec:comp}, but in freestyles. The high-quality alpha mattes are generated by specialists via image editing software, \eg, Adobe Photoshop and GIMP. We show some examples in Figure~\ref{fig:refmatte-rw100}. Furthermore, we show some statistics of RefMatte and RefMatte-RW100 in Table~\ref{tab:statistics}, including the number of images, alpha mattes, text descriptions, categories, and the average length of texts.
\footnote{More details of RefMatte, including the distribution of matting entities, linguistic details, and statistics are in the supplementary materials.}.

\begin{figure*}[t]
    \centering
    \includegraphics[width=\linewidth]{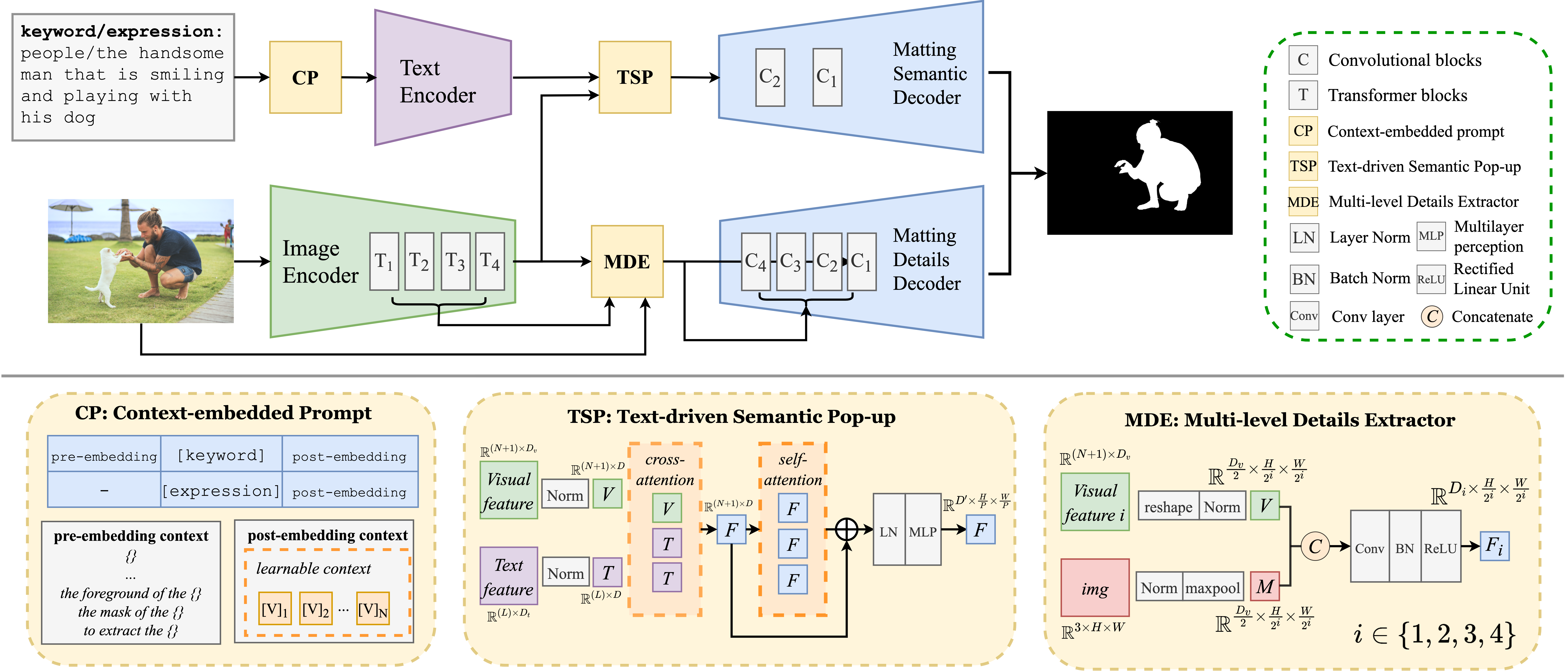}
    \caption{The diagram of the proposed method CLIPMat. The top indicates the whole pipeline, and the bottom describes each module.}
    \label{fig:clipmat}
\end{figure*}

\section{A Strong Baseline: CLIPMat}

\subsection{Overview}
Motivated by the success of large-scale pre-trained vision-language models like CLIP~\cite{clip} on downstream tasks, we also adopt the text encoder and image encoder from CLIP as our backbone. We choose ViT-B/16 and ViT-L/14~\cite{vit} as the image encoder backbone (to demonstrate that the scalability of model size also matters in image matting for the first time). As for the decoder, different from RIS methods~\cite{Rao2021DenseCLIPLD,clipseg} that predict a coarse segmentation mask through a single decoder, RIM is a task that requires both global semantic and local details information~\cite{gfm}. Thus, we utilize the dual-decoder framework from state-of-the-art matting methods~\cite{p3m,aim} to predict a trimap and the alpha matte in the transition area, respectively. We name them the matting semantic decoder and matting details decoder in CLIPMat. The input of our method is an image with a text description, which can be either a keyword (\eg \texttt{people}) or an expression (\texttt{the handsome man that is smiling and playing with his dog}), as shown in Figure~\ref{fig:clipmat}. The output is the meticulous alpha matte of the target object.

\subsection{CP: Context-embedded Prompt}

Although some previous works have already adopted prompt engineering~\cite{clip,coop} to enhance the understanding ability of the text input, how to adapt them in RIM is unexplored. In our work, we design two kinds of contexts to be embedded in the original prompt, named pre-embedding context and post-embedding context, as shown in Figure~\ref{fig:clipmat}. Both of them have been proven effective in the experiments. We present the details as follows.

\noindent\textbf{Pre-embedding context} For the keyword setting, to reduce the gap between a single word and the CLIP model pre-trained on long sentences, we create a bag of matting-related customized prefix context templates, including \textit{``the foreground of $\left\{ keyword \right\}$"}, \textit{``the mask of $\left\{ keyword \right\}$"}, \textit{``to extract the $\left\{ keyword \right\}$"} and so on. We add the pre-embedding context to the keyword directly before tokenization, ensuring that the text encoder can understand the image matting task by adapting the encoded knowledge during pre-training.

\noindent\textbf{Post-embedding context} To improve the ability of the text encoder to understand the text, we follow the work~\cite{coop} to add some learnable context appended to the tokenized text in both keyword and expression settings. Since the length of text space and context is different in the two settings, we use 14 and 69 for text length in keyword and expression settings, respectively, while the length of learnable context is fixed to 8 for both settings.

\begin{table*}[htbp]
\begin{center}
\caption{Results on the RefMatte test set in two settings and the RefMatte-RW100 test set.}
\label{tab:exp}
\resizebox{\linewidth}{!}{
\begin{tabular}{c|c|c|ccc|ccc|ccc}
\hline
 \multirow{2}{*}{Method}&\multirow{2}{*}{Backbone}&\multirow{2}{*}{Refiner} &\multicolumn{3}{c|}{Keyword-based setting} & \multicolumn{3}{c|}{Expression-based setting}& \multicolumn{3}{c}{RefMatte-RW100}\\
 \cline{4-12}
 & &&SAD & MSE & MAD & SAD & MSE & MAD & SAD & MSE & MAD \\
\hline
MDETR~\cite{mdetr} &  ResNet-101~\cite{he2016deep} & - & 32.27 &  0.0137& 0.0183 & 84.70 &0.0434 &0.0482&131.58 &0.0675 &0.0751\\
\hline
CLIPSeg~\cite{clipseg} & ViT-B/16~\cite{vit} & - & 17.75& 0.0064&0.0101 & 69.13& 0.0358& 0.0394& 211.86  &0.1178&0.1222\\
\hline
CLIPMat & ViT-B/16 & - & 9.91 & 0.0028 & 0.0057 & 47.97 & 0.0245 & 0.0273 & 110.66 & 0.0614 & 0.0636\\
\hline
CLIPMat & ViT-B/16 & yes & 9.13 & 0.0026 & 0.0052 & 46.38 & 0.0239 & 0.0264 & 107.81 & 0.0595 & 0.0620\\
\hline
CLIPMat & ViT-L/14~\cite{vit} & - & 8.51 & 0.0022 & 0.0049 & 42.05 & 0.0212 & 0.0238 & 88.52 & 0.0488 & 0.0510\\
\hline
CLIPMat & ViT-L/14 & yes &\textbf{8.29} & \textbf{0.0022} & \textbf{0.0027} & \textbf{40.37} & \textbf{0.0205} & \textbf{0.0229} & \textbf{85.83} & \textbf{0.0474} & \textbf{0.0495} \\
\hline
\end{tabular}
}
\end{center}
\end{table*}

\subsection{TSP: Text-driven Semantic Pop-up}

To ensure the text feature from the text encoder can provide better guidance on dense-level visual semantic perception, we propose a module named TSP (text-driven semantic pop-up) to process the text and visual features before the matting semantic decoder. Specifically, we abandon the last project layer in both the image encoder and text encoder to keep the original dimension. Thus, the input of TSP is the visual feature $x_v\in \mathbb{R}^{(N+1)\times D_v}$ and text feature $x_t\in \mathbb{R}^{L\times D_t}$, where $N=HW/{P^2}$ stands for the resulting number of patches after ViT transformer~\cite{vit}. On the other hand, $L$ stands for the total length of the text and embedding context, in our cases, which is 22 for the keyword-based setting and 79 for the expression-based setting. We first normalize them through layer norm, linear projection, and another layer norm to achieve the same dimension $D$. We then pop up the semantic information from the visual feature under the guidance of the text feature via cross-attention~\cite{attention}. In addition, we adopt self-attention to further refine the visual feature with a residual connection. Finally, we pass through the feature to layer norm and a multilayer perception, obtaining the feature of size $\mathbb{R}^{D'\times h \times w}$, where $h=\frac{H}{P}$ and $W=\frac{W}{P}$. The output feature is used as the input to the semantic decoder. Since it has already encoded high-level visual semantic information, we only use two convolution blocks to predict the trimap. Each contains two convolution layers and a bilinear upsampling layer with a stride 4. We adopt the cross-entropy loss in the semantic decoder following~\cite{gfm}.

\subsection{MDE: Multi-level Details Extractor}
Same as TSP, we also abandon the final projection layer from the CLIP image and text encoder. Since the matting detail decoder requires local detail information to generate meticulous alpha matte, we design the MDE to extract useful local details from both the original image and multi-level features from the image encoder. Specifically, we take the output features from all four transformer blocks in the CLIP image encoder, denoted as $x_v^i$ where $i\in\left\{ 1,2,3,4 \right\}$. For each $x_v^i$, we pass it and the original image $X_m$ to MDE. For $x_v^i$, we first reshape and then normalize it by a $1\times 1$ convolution layer. For $X_m$, we first normalize it by a $1\times 1$ convolution layer and then down-sample it to the same size as $x_v^i$ via max pooling. They are concatenated to form $x_f^i$ and fed into a convolution layer, a batch norm layer, and a ReLU activation layer. Finally, the output feature is used as the input to the corresponding decoder layer at each level via a residual connection. Following~\cite{gfm}, we use the alpha loss and Laplacian loss in the matting details decoder. The outputs from the two decoders are merged through the collaboration module~\cite{gfm} to get the final output, supervised by the alpha loss and Laplacian loss. More details of the method can be found in the supplementary materials. 
 
\section{Experiments}

\subsection{Experiment Settings}
\label{sec:implementaion}
Since there are no prior methods designed for the new RIM task, we choose state-of-art methods from relevant tasks, \ie, CLIPSeg~\cite{clipseg} and MDETR~\cite{mdetr}, which are two representative methods for the RIS and RES tasks, for benchmarking. All the methods, and CLIPMat are trained on the RefMatte training set and evaluated in two settings, \ie, the keyword-based setting and expression-based setting.

\noindent\textbf{Implementation details} We resize the image to $512\times512$ and adopt data augmentation following~\cite{gfm} to reduce the domain gap of composite images. We use the Adam optimizer. We train CLIPMat on two NVIDIA A100 GPUs with the learning rate fixed to 1e-4. For the ViT/B-16 backbone, the batch size is 12 and is trained for 50 epochs (about 1 day). For the ViT/L-14 backbone, the batch size is 4 and is trained for 50 epochs (about 3 days). For CLIPSeg~\cite{clipseg} and MDETR~\cite{mdetr}, we use the code and the weights pre-trained on VGPhraseCut~\cite{phrasecut} provided by the authors for training them. However, we have not pre-trained CLIPMat on VGPhraseCut since we find that directly training it on RefMatte could provide better performance.

\noindent\textbf{Evaluation metrics} Following the common practice in previous matting methods~\cite{dim,gfm,p3m}, we use the sum of absolute differences (SAD), mean squared error (MSE), and mean absolute difference (MAD) as evaluation metrics, which are averaged over all the entities in the test set. 

\noindent\textbf{Matting refiner} To further improve the details of alpha matte, we propose a coarse map-based matting method as an optional post-refiner. Specifically, we modify P3M~\cite{p3m} to receive the original image and the predicted alpha matte as input and train it on RefMatte to refine the alpha matte.

\begin{figure*}[t]
    \centering
    \includegraphics[width=\linewidth]{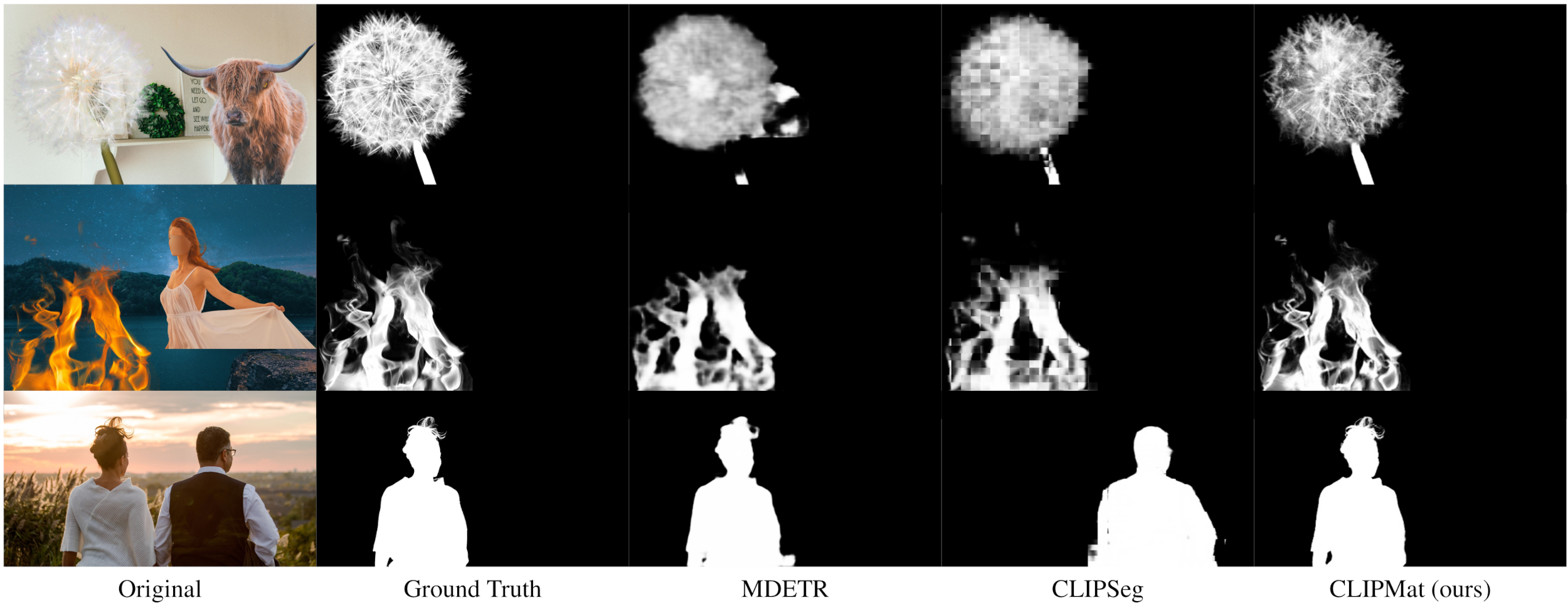}
    \caption{Subjective comparison of different methods on RefMatte and RefMatte-RW100 in different settings. The text inputs from the top to the bottom are: 1) \textit{dandelion}; 2) \textit{the flame which is lightsalmon and non-salient}; 3) \textit{the woman who is with her back to the camera}.}
    \label{fig:exp}
\end{figure*}

\subsection{Main Results}

\subsubsection{Keyword-based Setting}
We evaluate MDETR~\cite{mdetr}, CLIPSeg~\cite{clipseg}, and CLIPMat on the keyword-based setting of the RefMatte test set, and show the quantitative results in Table~\ref{tab:exp}. As can be seen, CLIPMat outperforms MDETR and CLIPSeg by a large margin using either the ViT-B/16 or ViT-L/14 backbone, validating the superiority of the proposed baseline method. Besides, we also show that using a larger backbone and the refiner could deliver better results. The best CLIPMat model reduces error of MDETR by about 75\% and the error of CLIPSeg by about 50\%, owing to the special design of the three modules. As seen from the top row in Figure~\ref{fig:exp}, with the input keyword \textit{dandelion}, CLIPMat is able to extract the very fine details of the target from the background with a similar color. However, both CLIPSeg and MDETR fail in this case, producing incomplete and blurry alpha mattes.

\subsubsection{Expression-based Setting}
We also evaluate these models on the RefMatte test set and RefMatte-RW100 under the expression setting. Similar to the keyword-based setting, the results in Table~\ref{tab:exp} also demonstrate the superiority of CLIPMat over MDETR and CLIPSeg, \eg, the best CLIPMat model reduces the error of MDETR on the RefMatte test set by over 50\% and the error of CLIPSeg on RefMatte-RW100 by about 60\%. Again, using a larger backbone and the refiner help reduce the error. As seen from the second row in Figure~\ref{fig:exp}, CLIPMat outperforms others in extracting the fine details of the flame, which are very close to the ground truth. The test image in the third row is from RefMatte-RW100. Compared with CLIPSeg, which produces the wrong foreground, CLIPMat is able to find the right foreground by pop-upping the correct visual semantic feature owing to the TSP module. The MDE module helps CLIPMat preserve more details, \eg, the woman's hair, compared with MDETR. The results show the good generalization ability of CLIPMat on real-world images and confirm the value of the proposed RefMatte dataset.

\begin{table}[htbp]
\begin{center}
\resizebox{\linewidth}{!}{
\begin{tabular}{cccc|ccc}
\hline
TSP & MDE & Pre-CP & Post-CP   & SAD & MSE & MAD \\
\hline
 & && & 22.88 & 0.0097 & 0.0131 \\
\checkmark& &  & & 18.28 & 0.0068 & 0.0105 \\
\checkmark& \checkmark& & & 14.55 & 0.0050 & 0.0083 \\
\checkmark& \checkmark&\checkmark & & 11.48 & 0.0036 & 0.0065 \\
\checkmark& \checkmark& &\checkmark & 12.96 & 0.0045 & 0.0074 \\
\checkmark& \checkmark& \checkmark&\checkmark & \textbf{9.91} & \textbf{0.0028} & \textbf{0.0057} \\
 \hline
\end{tabular}
}
\end{center}
\caption{Ablation studies results. TSP: text-driven semantic pop-up; MDE: multi-level details extractor; Pre-/Post-CP: pre or post context-embedded prompt. We use ViT-B/16 as the backbone.}
\label{tab:ablation}
\end{table}

\subsection{Ablation Studies}
We conduct ablation studies to validate the effectiveness of our proposed modules. The experiments are carried out in the keyword-based setting of RefMatte. We show the results in Table~\ref{tab:ablation}. We can see that each module contributes to performance improvement in terms of all the metrics, \eg, the combination of MDE and TSP reduces the SAD from 22.88 to 14.55. The use of CP further reduces the SAD to 9.91, validating that the customized matting prefix and the learnable queries provide useful context for the text encoder to understand the language instruction for image matting.\footnote{We show more ablation studies, experiment details, failure cases, and more visual results in the supplementary materials.}

\section{Conclusion}

In this paper, we define a novel task named referring image matting (RIM), establish a large-scale dataset RefMatte, and provide a baseline method CLIPMat. RefMatte provides a suitable test bed for the study of RIM, thanks to its large scale, high-quality images, and abundant annotations, as well as two well-defined experiment settings. Together with the RefMatte-RW100, they can be used for both in-domain and out-of-domain generalization evaluation. Besides, the CLIPMat shows the value of special designs for the RIM task and serves as a valuable reference to the model design. We hope this study could provide useful insights to the image matting community and inspire more follow-up research. 

\clearpage
{\small
\bibliographystyle{ieee_fullname}
\bibliography{matting}
}

\end{document}


\title{Referring Image Matting\\ (Supplementary Material) }
\author{Jizhizi Li, Jing Zhang, and Dacheng Tao\\
The University of Sydney, Sydney, Australia\\
{\tt\small jili8515@uni.sydney.edu.au, jing.zhang1@sydney.edu.au, dacheng.tao@gmail.com}
}

\maketitle

\section{The Significance of RIM}

\begin{figure*}[hbp]
    \centering
    \includegraphics[width=\linewidth]{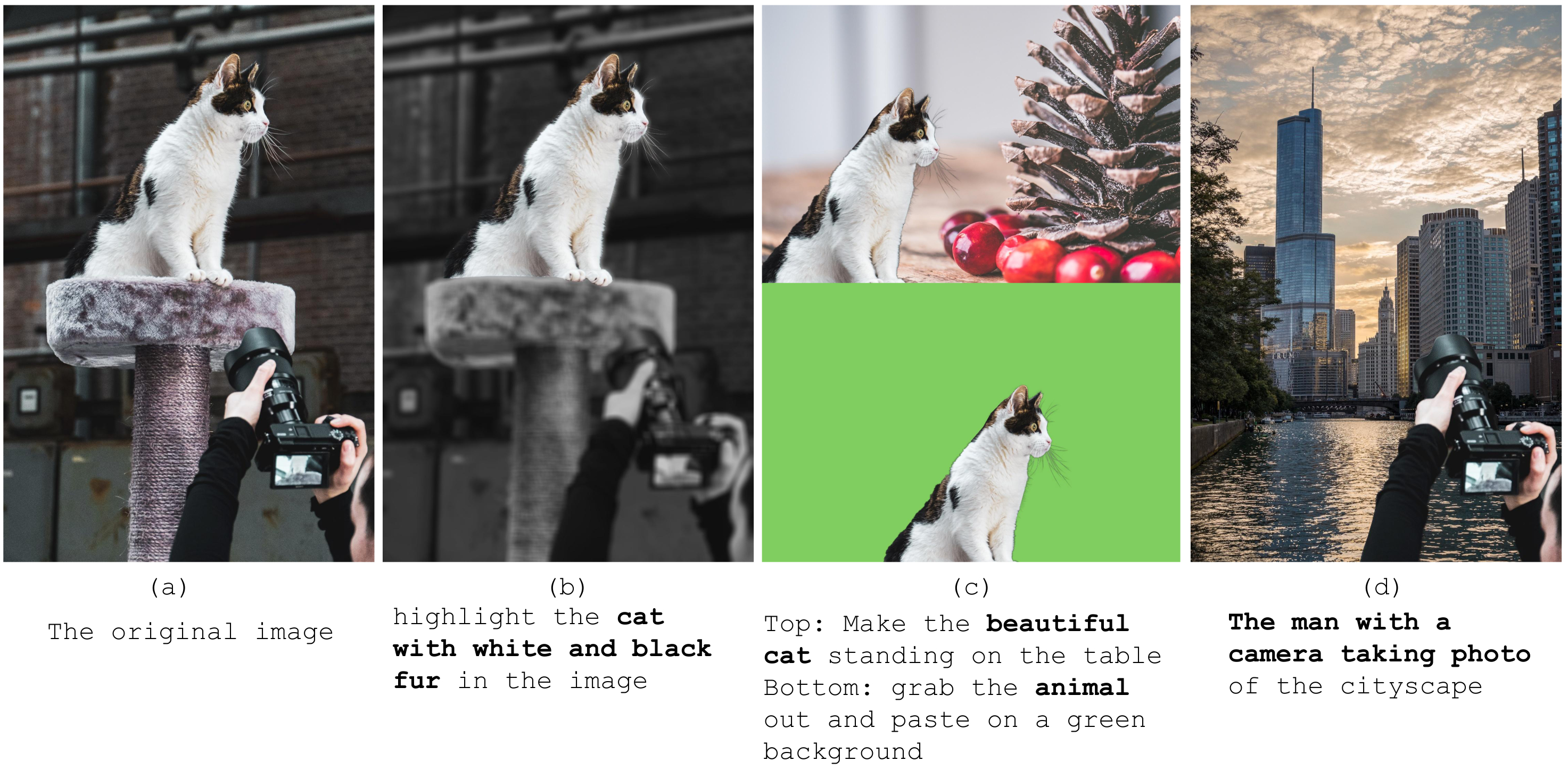}
    \caption{A case study of utilizing RIM for interactive image editing. The text inputs guide the interactive editing while the bold font indicates the given language descriptions.}
    \label{fig:case_study}
\end{figure*}

Referring Image Matting (RIM) refers to extracting the soft and accurate foregrounds from the images based on given language descriptions. As a new task, we discuss the significance of RIM from the aspects of the industry impacts and academic impacts.

RIM, as a dense prediction problem that needs the collaboration of language and vision, provides a pathway to controllable image matting. Compared with conventional automatic image matting methods that can only predict a fixed set of pre-defined categories, e.g, people~\cite{p3m}, animal~\cite{gfm}, classes~\cite{aim} or all the salient objects at one time~\cite{hatt, lf}, RIM is able to predict the alpha matte of the specific foregrounds under language guidance, serving as a fundamental task for various downstream applications such as interactive image editing, human-machine interaction, virtual reality, augmented reality, film production, e-commerce promotion, etc. Compared with the other auxiliary user input-based image matting methods, e.g., trimap-based~\cite{dim} and scribble-based~\cite{levin2007closed}, RIM provides much more freedom for the users as the inputs are simple and straightforward language descriptions. 

As a novel task that has not been explored yet, RIM opens up lots of new research directions in the area of image matting. For example, how to align the visual and text features to better exploit both semantic and details information for such a dense level problem, how to bridge the domain gap between the typical composite training data and real-world testing data, how to distinguish between the same type objects, etc. Our proposed RefMatte, along with the baseline method CLIPMat, can serve as an initial starting point for such studies. Specifically, we discuss a case study and the differences with a related task Referring Image Segmentation (RIS) as follows to further emphasize the significance of RIM.

\noindent\textbf{A case study of RIM.} Here we provide a case study of RIM as a concrete example of utilizing RIM in downstream applications, e.g., interactive image editing. As can be seen from Figure~\ref{fig:case_study}, RIM is able to provide various interactive image editing results based on customized user text input, including highlighting any objects of interest and pasting the objects of interest to a reasonable background or a pure color. Different from the previous image matting methods that can only extract fixed-type foregrounds or all the salient foregrounds, RIM can easily separate the adherent foregrounds and focus on the one that best matches the language description. In this way, users can flexibly perform post-processing on either the \textit{cat with white and black fur} or \textit{the man with a camera taking photo} or any other customized language they like. 

\noindent\textbf{The differences with RIS.} RIM is a very different task compared with RIS from the following several aspects:
\begin{enumerate}
  \item RIS is only able to predict the coarse contour shape of the foreground from a low-resolution image, while RIM can predict the very fine details of the foregrounds from a high-resolution image.
  \item RIS is a binary classification problem, while RIM is a pixel-level regression problem, requiring the datasets and methods to be very different from the two problems.
  \item Datasets designed for RIS~\cite{phrasecut, Krishna2016VisualGC} usually have low-resolution images and objects with very coarse shapes, while datasets for RIM should be all high-resolution images that preserve as many details as possible, as in our RefMatte.
  \item Methods designed for RIS usually focus on extracting semantic features~\cite{clipseg, mdetr} while RIM methods require features at both the (global) semantic level and (local) details level, making it more challenging, as in our proposed CLIPMat.
\end{enumerate}

\section{More Details about RefMatte}
In this section, we provide more details about RefMatte, including the distribution of matting entities, linguistic details, and more statistics and visual results.

\subsection{The Distribution of Matting Entities}
\label{sec:matting_datasets} 

\begin{table*}[htbp]
\begin{center}
\begin{tabular}{c|c|c|ccc|cc}
\hline
Dataset & Category & Split & \makecell[c]{\#Entities} &\makecell[c]{\#Categories}& \makecell[c]{\#Attrs.\\per Entity}& \makecell[c]{\#Entities in \\RefMatte train} & \makecell[c]{\#Entities in \\RefMatte test}   \\
\hline
  \multirow{2}{*}{AM-2k~\cite{gfm}}& \multirow{2}{*}{animal}& train & 1800& 20 & \multirow{2}{*}{3}& 1800&- \\
  & &test &200  & 20& & -&200 \\
  \cline{1-8}
 \multirow{3}{*}{P3M-10k~\cite{p3m}}& \multirow{3}{*}{human}& train & 9186& 1& \multirow{3}{*}{6}& 9186&- \\
 &&test-1& 485& 1& & - &485 \\
 &&test-2&492 & 1& & - &492 \\
 \cline{1-8}
 \multirow{1}{*}{AIM-500~\cite{aim}}& \multirow{1}{*}{objects}&test& 200& 93& \multirow{1}{*}{3}& 95&105 \\
 \cline{1-8}
 \multirow{2}{*}{SIM~\cite{sim}}& \multirow{2}{*}{objects}&train& 271& 82& \multirow{2}{*}{3}&271 &- \\
 &&test& 41& 27 & &2 &39 \\
 \cline{1-8}
 \multirow{2}{*}{DIM~\cite{dim}}& \multirow{2}{*}{objects}&train& 224&75& \multirow{2}{*}{3}&224 &-  \\
 &&test&38 &27 & & 7&31  \\
 \cline{1-8}
 \multirow{2}{*}{HATT~\cite{hatt}}& \multirow{2}{*}{objects}&train& 210&58& \multirow{2}{*}{3}& 210&-  \\
 &&test&40 &30 & &4 &36  \\
 \hline
 \multirow{2}{*}{RefMatte (ours)}&\multirow{2}{*}{all-types}&train&11799 &230 & \multirow{2}{*}{3/6}& 11700&-  \\
 &&test& 1388&66 & &- & 1388 \\
 \hline
\end{tabular}
\end{center}
\caption{Statistics of the matting entities in our RefMatte which come from previous matting datasets.}
\label{tab:matting_datasets}
\end{table*}

Here, we present more details about the previous matting datasets that we used to retrieve the entities in our RefMatte, including the number of entities, categories, and attributes we generate per entity and the proportion of each dataset in our RefMatte train and test split. The results can be seen from Table~\ref{tab:matting_datasets}. As can be seen, AM-2k~\cite{gfm} contributes all images of the animal categories, P3M-10k~\cite{p3m} contributes all images of the human category, and the others contribute the objects types like smog, plant, transparent glasses, etc. The number of attributes per entity is 3 for animals and objects, and 6 for humans. The proportion of each dataset in regard to the train and test split is shown in the table too. As can be seen, we try to reserve the original split in each matting dataset except for migrating the long-tailed categories to the RefMatte train set. However, the distribution of entities is unbalanced since most of them are human or animal. Thus, we duplicate some entities to form a balanced proportion of \textit{human, animals, and objects} as \textit{5:1:1}. The details are in the main paper.

\subsection{Linguistic Details in RefMatte}
In this subsection, we provide the linguistic details we used for constructing RefMatte, including the candidate words for the attributes and synonyms of the human type, the definitions of transparent and salient entities, syntax templates in the basic expressions, and the relationship words in the absolute/relative position expressions.

\textbf{Candidate words for the human type} To generate more precise synonyms, we define the basic synonyms for human type as \texttt{human being, citizenry, person, individual, mankind, mortal}. In addition to them, depending on the age, age group, and gender we defined, we provide more candidate words to serve as reasonable synonyms, which are shown in Table~\ref{tab:synonyms_human}. All the candidate words are used to form the expression randomly.

\begin{table}[htbp]
\begin{center}
\begin{tabular}{c|c|c|c}
\hline
Age & Age Group & Synonyms for Female & Synonyms for Male \\
\hline
0-2 & \multirow{3}{*}{child} &\multirow{3}{*}{\texttt{\makecell[c]{baby girl, little girl,\\ girl}}}& \multirow{3}{*}{\texttt{\makecell[c]{baby boy, little boy, \\boy}}} \\
4-6 & & & \\
8-12 &  & & \\
\hline
15-20 & youth & \texttt{\makecell[c]{girl, teenager, adolescent,\\ miss, missy, \\young lady, young woman}}
& \texttt{boy, teenager, adolescent}\\
\hline
25-32 & \multirow{3}{*}{adult} & \multirow{3}{*}{\texttt{woman,lady}}&\multirow{3}{*}{\texttt{man}} \\
38-43 & & & \\
48-53 &  & & \\
\hline
60-100 & senior & \texttt{\makecell[c]{old woman, senior citizen, \\pensioner}}& \texttt{\makecell[c]{old man, senior citizen,\\ pensioner}} \\
 \hline
\end{tabular}
\end{center}
\caption{Synonyms of female and male at different age groups.}
\label{tab:synonyms_human}
\end{table}

\textbf{Definitions of transparent and salient entities} Following Li et al.~\cite{aim}, we add the attributes \texttt{transparent} and \texttt{salient} for all the entities in RefMatte. We define the entities with category name synonyms including~\textit{smoke, glass, water, gauze, lace, ice, bubble wrap, plastic bag, net, fire, flame, cloth, mesh bag, mesh, wine glass, ice cube, spider web, wine, cloud smog, veil, wedding dress, fishing net, cloth net, light, water drop, drip, dew, crystal stone, beer} as~\texttt{transparent} ones. Those entities with synonyms like~\textit{smoke, water, gauze, lace, fire, flame, net, leaves, spider web, mesh, wine, smog, light, water spray} as~\texttt{non-salient} ones. For all the other entities, we can easily define them as~\texttt{non-transparent} ones or~\texttt{salient} ones, e.g., \textit{human} and \textit{animal} are both \texttt{salient} and \texttt{non-transparent}.

\textbf{Syntax templates in the basic expressions} We generate the basic expressions following the syntax templates as shown in Table~\ref{tab:synonyms_be}, the templates are different for the human type and others since a human has six attributes and others have only three.

\begin{table}[htbp]
\begin{center}
\begin{tabular}{c|c|c}
\hline
type & attributes  & syntax template \\
\hline
human & \texttt{\makecell[l]{$<att_0>$: gender\\$<att_1>$: age\\$<att_2>$: non-transparent\\$<att_3>$: salient \\$<att_4>$: color \\$<att_5>$: clothes type }}& \texttt{\makecell[l]{the $<att_{0-3}>$ $<obj_0>$ with the $<att_{4-5}>$\\the $<att_{0-3}>$ $<obj_0>$ wearing the $<att_{4-5}>$\\the $<att_{0-3}>$ $<obj_0>$ in the $<att_{4-5}>$ \\ the $<att_{0-3}>$ $<obj_0>$ who is dressed in $<att_{4-5}>$}}\\
\hline
others & \texttt{\makecell[l]{$<att_0>$: color \\$<att_1>$: non-/ transparent\\$<att_2>$: non-/ salient}}& \texttt{\makecell[l]{the $<att_{0-2}>$ $<obj_0>$\\the $<obj_0>$ which is $<att_{0-2}>$}}\\
\hline
\end{tabular}
\end{center}
\caption{Syntax template in the basic expression.}
\label{tab:synonyms_be}
\end{table}

\textbf{Relationship words in the absolute/relative position expressions} As discussed in the paper, the syntax templates for the absolute position expressions are \texttt{the/a <$att_0$> <$att_1$>...<$obj_0$> <$rel_0$> the photo/image/picture} or \texttt{the/a <$obj_0$> which/that is <$att_0$> <$att_1$> <$rel_0$> the photo/image/picture}. The syntax templates for the relative position expression are \texttt{the/a <$att_0$> <$att_1$>...<$obj_0$> <$rel_0$> the/a <$att_2$> <$att_3$>...<$obj_1$>} or \texttt{the/a <$obj_0$> which/that is <$att_0$> <$att_1$> <$rel_0$> the/a <$obj_1$> which/that is <$att_2$> <$att_3$>}. Here we provide the candidate prepositional phrases for the relationship words \texttt{<$rel_0$>} in Table~\ref{tab:relation_aperpe} for each position relationship. Please note that the relationship~\texttt{middle} is only used in the absolute position expressions.

\begin{table}[htbp]
\begin{center}
\begin{tabular}{c|c|c}
\hline
\makecell[c]{Position\\Relationship} & \makecell[c]{$<rel_0>$ in \\absolute position expression} & \makecell[c]{$<rel_0>$ in \\relative position expression} \\
\hline
left& \texttt{\makecell[c]{at the most left side of,\\on the far left of,\\at the leftmost edge of,\\farthest to the left of}} & \texttt{\makecell[c]{to the left of,\\on the left side of,\\at the left side of,beside,\\next to,close to,near}}\\
\hline
right &\texttt{\makecell[c]{at the most right side of,\\on the far right of,\\at the rightmost edge of,\\farthest to the right of}}&\texttt{\makecell[c]{to the right of,\\on the right side of,\\at the right side of,beside,\\next to,close to,near}} \\
\hline
middle   &\texttt{\makecell[c]{in the middle of,\\n the center of}}& -\\
\hline
top & \texttt{\makecell[c]{on top of,\\in the upper part of}}& \texttt{\makecell[c]{above,over,on top of,on}}\\
\hline
bottom & \texttt{\makecell[c]{below,in the lower part of}}& \texttt{\makecell[c]{below,under,underneath}}\\
\hline
in front of  & \texttt{\makecell[c]{in front of}}&\texttt{\makecell[c]{in front of}}\\
\hline
behind & \texttt{\makecell[c]{behind,in the back of,\\at the back of}}& \texttt{\makecell[c]{behind,in the back of,\\at the back of}}\\
\hline
\end{tabular}
\end{center}
\caption{Relationship words in the absolute/relative position expressions.}
\label{tab:relation_aperpe}
\end{table}

\subsection{The Statistics of RefMatte}
We present more details about the statistics of RefMatte in Table~\ref{tab:statistics}. For keyword-setting, since the text description is the entry-level category name, we remove the images with multiple entities belonging to the same category to avoid semantic ambiguity. Consequently, we have 30,391 images in the training set and 1,602 images in the test set in this setting. The numbers of alpha mattes, text descriptions, categories, attributes, and relationships are shown in the following columns, respectively. The average text length in the keyword-based setting is about 1, since there is usually a single word for each category, while it is much larger in the expression-based setting, i.e., about 17 in RefMatte and 12 in RefMatte-RW100. 

\begin{table}[htbp]
\begin{center}
\begin{tabular}{c|c|ccc|c|cc|cc|c}
\hline
Dataset  &Split & \makecell[c]{Image \\Num.} &\makecell[c]{Matte \\Num.}& \makecell[c]{Text \\Num.} &\makecell[c]{Category \\Num.} & \makecell[c]{Attrs.\\Num.} &\makecell[c]{Rels.\\Num.} & \makecell[c]{Text \\Length} \\
\hline
  RefMatte & train & 30,391 & 77,849 & 77,849 & 230 & - & - & 1.06\\
  Keyword&  test & 1,602 & 4,085 & 4,085 & 66 & - & - & 1.04\\
  \hline
  RefMatte & train & 45,000 & 112,506 & 449,624 & 230  & 132&31&16.86\\
  Expression & test & 2,500 & 6,243 & 24,972 & 66 & 102&31&16.80 \\
 \cline{1-9}
 RefMatte-RW100 & test & 100 & 221 & 884 & 29 &135&34& 12.01\\
 \hline
\end{tabular}
\end{center}
\caption{Statistics of RefMatte and RefMatte-RW100 regarding to the number of images, alpha mattes, text descriptions, categories, attributes, relationship words, and the average length of texts.}
\label{tab:statistics}
\end{table}

\begin{figure*}[htbp]
\centering
\captionsetup[subfloat]{labelformat=empty,justification=centering}
\subfloat[(a)]{\includegraphics[width=.33\linewidth]{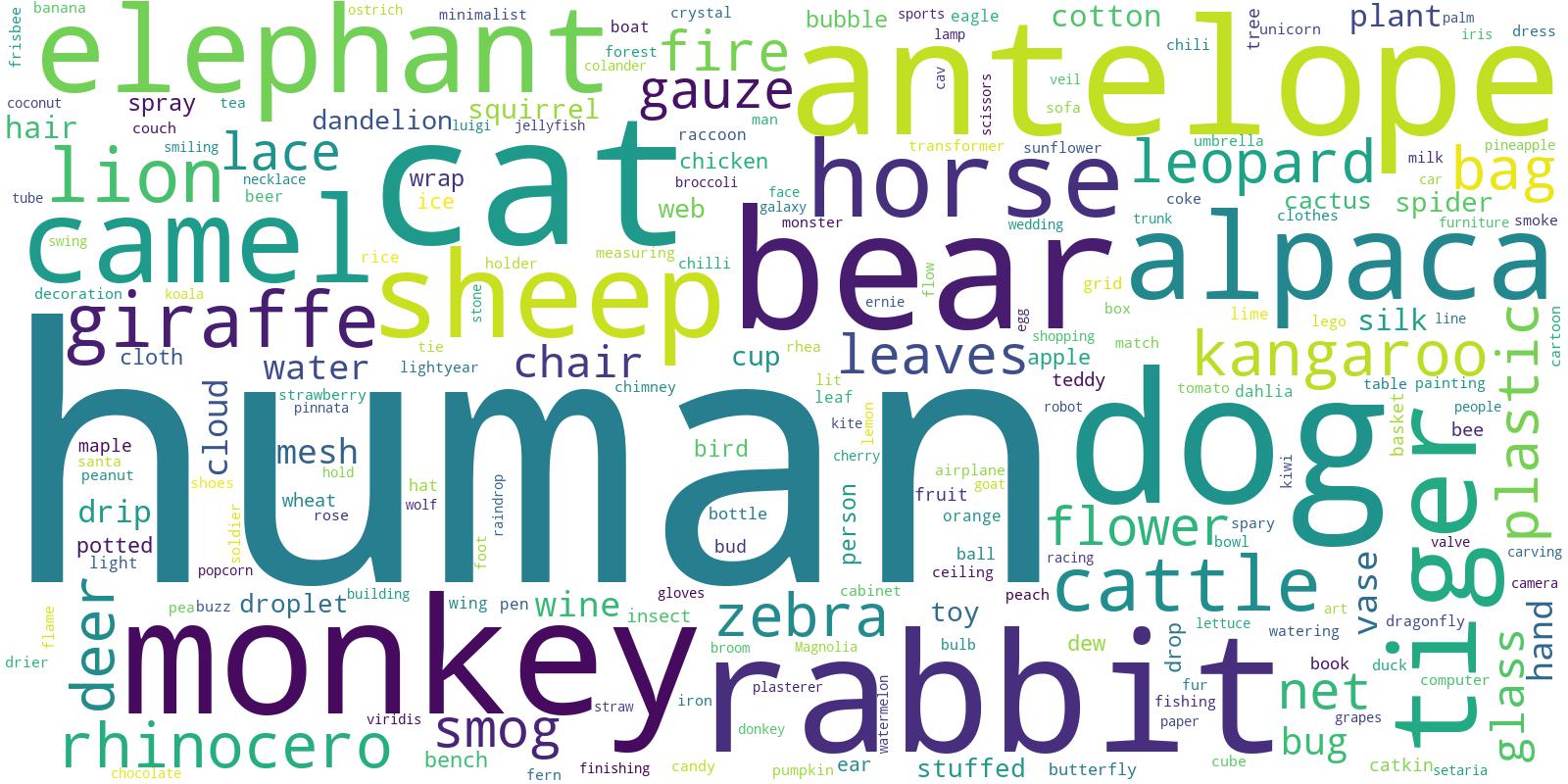}}
\subfloat[(b)]{\includegraphics[width=.33\linewidth]{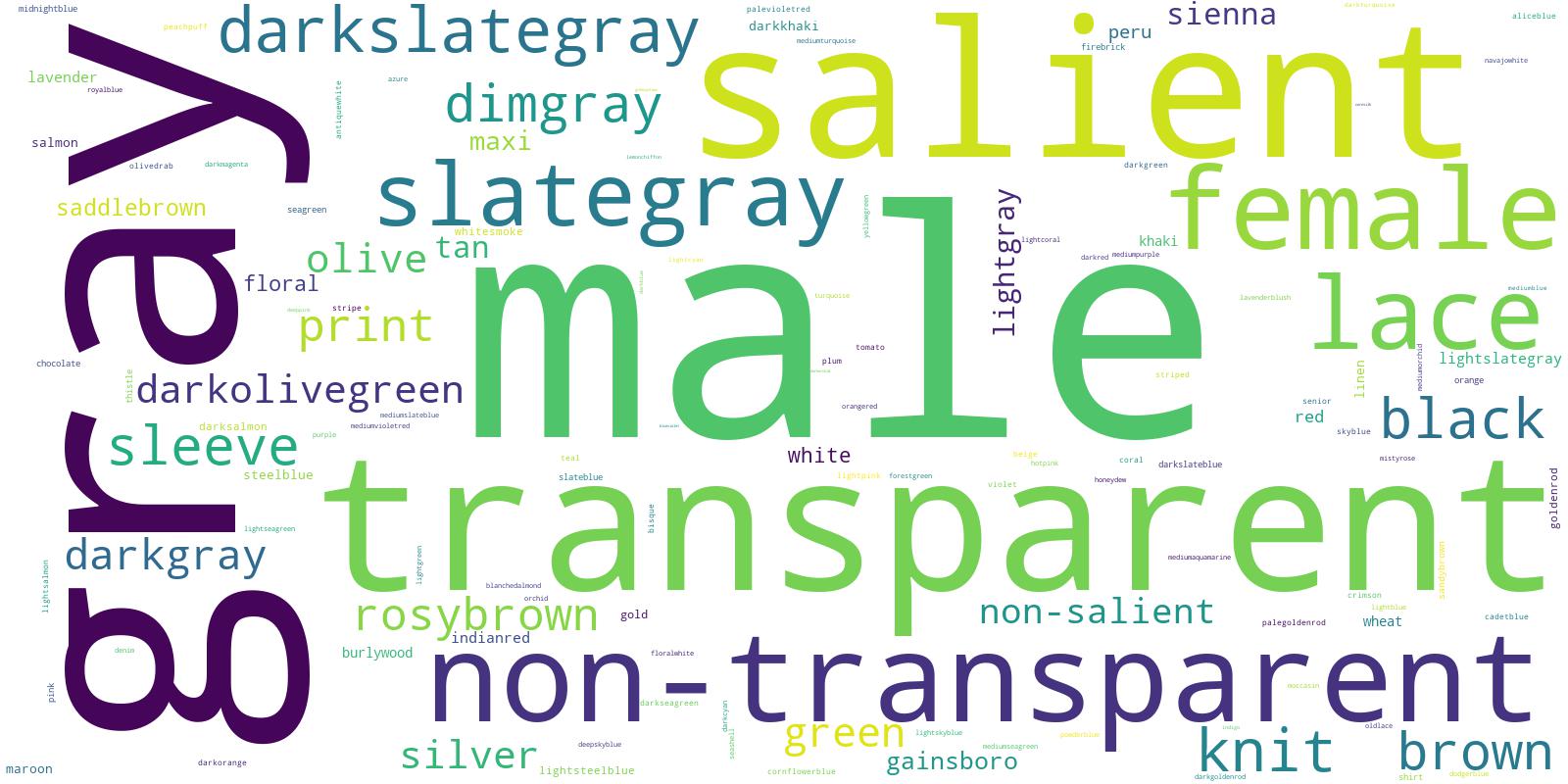}}
\subfloat[(c)]{\includegraphics[width=.33\linewidth]{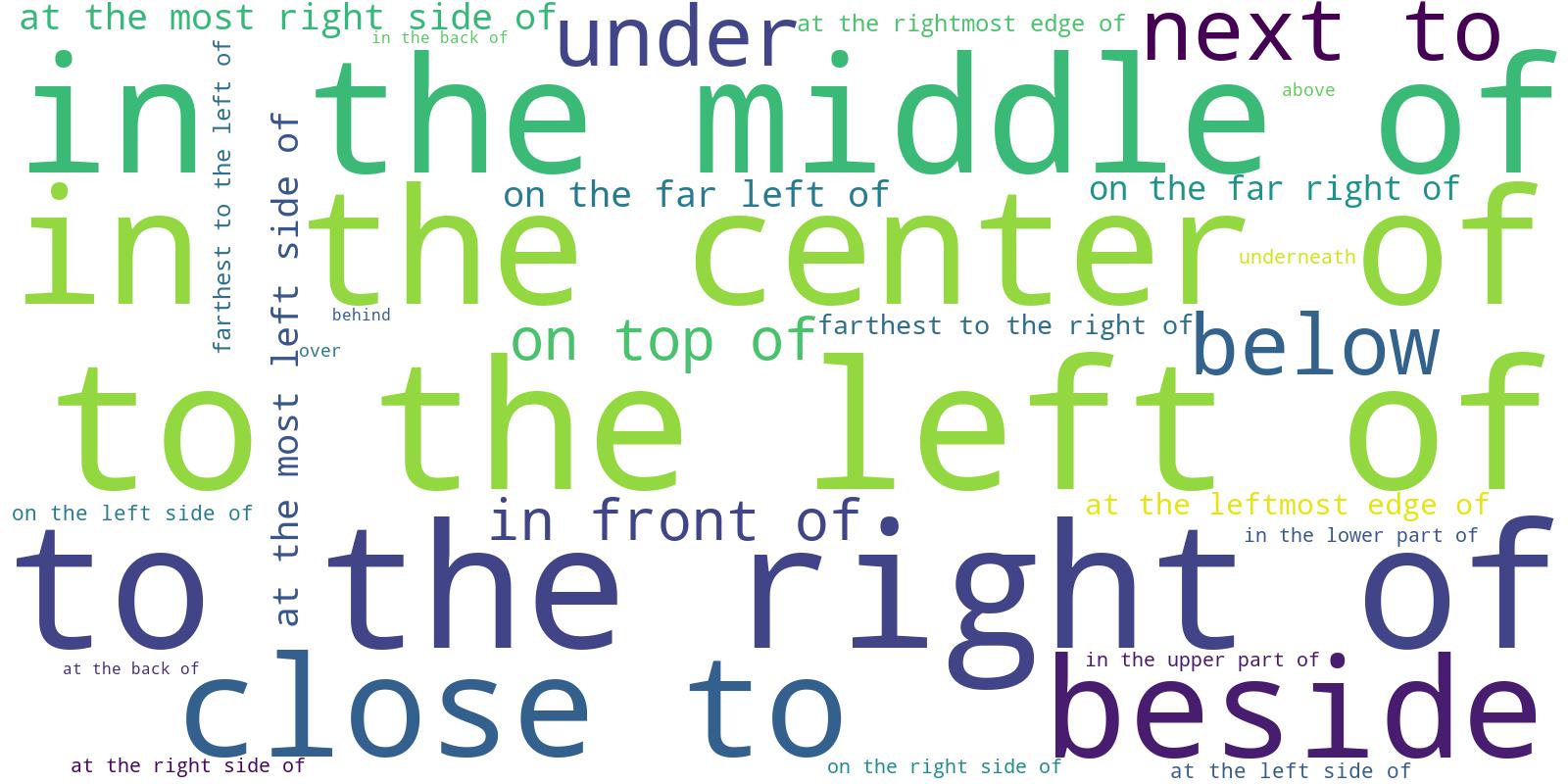}}
\caption{The word cloud of the keywords (a), attributes (b), and relationships (c) in RefMatte.}
\vspace{-0.5\baselineskip}
\label{fig:statistics}
\end{figure*}

We also generate the word cloud of the keywords and attributes of the entities as well as the relationships between the entities in RefMatte in Figure~\ref{fig:statistics}. As can be seen, the dataset has a large portion of humans and animals since they are much more common in the image matting task. The most frequent attributes in RefMatte are \textit{male, gray, transparent, and salient}, while the relationship words are more balanced, containing all kinds of relationships.

\subsection{More Examples of RefMatte}
We show more examples from our RefMatte training set and test set including their composition relations, keywords, basic expressions, absolute position expressions, and relative position expressions in Figure~\ref{fig:refmatte_supp}. We also show more examples from our RefMatte-RW100 test set, including their basic expression, absolute position expressions, relative position expressions, and free expressions in Figure~\ref{fig:refmatte_rw}. The green dots in both figures indicate the target objects.

\section{More Details of CLIPMat}

In this section, we present more details of our proposed baseline method CLIPMat, including more details about the three modules, \ie, CP (context-embedded prompt), TSP (text-driven semantic pop-up), and MDE (multi-level details extractor). We also show the network structure of CLIPMat in Table~\ref{Tab:clipmat_structure}.

\noindent\textbf{Matting-related prefix templates} We use a bag of words to serve as the matting-related prefix templates, aiming to reduce the gap between the long sentence used during pre-training CLIP~\cite{clip} and the ``single'' word in the keyword-setting in RefMatte. Specifically, the templates in the bags of words are \textit{``$\left\{ keyword \right\}$"}, \textit{``a photo of a $\left\{ keyword \right\}$"}, \textit{``a photograph of a $\left\{ keyword \right\}$"}, \textit{``an image of a $\left\{ keyword \right\}$"}, \textit{``a photo of the $\left\{ keyword \right\}$"},  \textit{``the foreground of the $\left\{ keyword \right\}$"}, \textit{``the mask of the $\left\{ keyword \right\}$"}, \textit{``the alpha matte of the $\left\{ keyword \right\}$"}, \textit{``to extract the $\left\{ keyword \right\}$"}. The experiments have proved the effectiveness of using them in enhancing the ability of the pre-trained CLIP text encoder for the image matting task.

\noindent\textbf{TSP details} With the input of TSP as visual feature from CLIP image encoder $x_v\in \mathbb{R}^{(N+1)\times D_v}$ and text feature from CLIP text encoder $x_t\in \mathbb{R}^{L\times D_t}$, where $N=HW/{P^2}$ stands for the number of patches (tokens) in the ViT transformer~\cite{vit}, the additional one dimension denotes the class token which is not involved during feature reshaping, and $L$ stands for the total length of the text and embedding context. We show the details of TSP as follows. First, both $x_v$ and $x_t$ pass through a layer norm, a linear layer, and another layer norm to align the feature dimension as $D$. Thus we have $x_v'$ and $x_t'$ with the same dimension $D$. We then pop up the semantic information from the visual feature by guiding it with the text feature through the cross-attention mechanism in transformer~\cite{attention}, thus we have $x_f\in \mathbb{R}^{(N+1)\times D}$. Furthermore, we adopt a self-attention mechanism to refine $x_f$ and we adopt the residual connection to ease optimization. Finally, we pass $x_f$ through a layer norm and a multilayer perception and then reshape it to $\mathbb{R}^{D'\times h \times w}$, where $h=\frac{H}{P}$ and $W=\frac{W}{P}$. This process can be formulated as follows:
\begin{gather}
x_v' = LN(Linear(LN(x_v))),\quad\quad x_v'\in \mathbb{R}^{(N+1)\times D}, \\
x_t' = LN(Linear(LN(x_t))),\quad\quad x_t\in \mathbb{R}^{L\times D}, \\
x_f = corss\_attn(x_v', x_t', x_t'),\quad\quad x_f\in \mathbb{R}^{(N+1)\times D}, \\
x_f = x_f+self\_attn(x_f, x_f, x_f),\quad\quad x_f\in \mathbb{R}^{(N+1)\times D}, \\
x_f = reshape(MLP(LN(x_f))),\quad\quad x_f\in \mathbb{R}^{D'\times h \times w}.
\end{gather}

\noindent\textbf{MDE details} For MDE, the input feature is one of the four transformer blocks in the CLIP image encoder and the original image, denoted as $x_v^i$ where $i\in\left\{ 1,2,3,4 \right\}$ and $X_m \in \mathbb{R}^{3\times H\times W}$, respectively. We show the details of MDE as follows. First, we reshape $x_v^i$ and then normalize it by a $1\times 1$ convolution layer, resulting in $x_v^i\in \mathbb{R}^{\frac{D_v}{2}\times \frac{H}{2^i} \times \frac{W}{2^i}}$. For $X_m$, we first normalize it by a $1\times 1$ convolution layer and then down-sample it to the same size as $x_v^i$ via max pooling, resulting in $x_m\in \mathbb{R}^{\frac{D_v}{2}\times \frac{H}{2^i} \times \frac{W}{2^i}}$. We then concatenate $x_v^i$ and $x_m$ to form $x_f^i\in \mathbb{R}^{D_v\times \frac{H}{2^i} \times \frac{W}{2^i}}$, which will be fed into a convolution layer, a batch norm layer, and a ReLU activation layer, results in the final output $x_f^i\in \mathbb{R}^{D_i\times \frac{H}{2^i} \times \frac{W}{2^i}}$. Finally, $x_f^i$ is used as the input to the corresponding decoder layer at each level via a residual connection, which can preserve the details. This process can be formulated as follows:
\begin{gather}
x_v^i = norm(reshape(x_v^i)),\quad\quad x_v^i\in \mathbb{R}^{\frac{D_v}{2}\times \frac{H}{2^i} \times \frac{W}{2^i}}, \\
x_m = maxpool(norm(x_m)),\quad\quad x_m\in \mathbb{R}^{\frac{D_v}{2}\times \frac{H}{2^i} \times \frac{W}{2^i}}, \\
x_f^i = concat(x_v^i,x_m),\quad\quad x_f^i\in \mathbb{R}^{D_v\times \frac{H}{2^i} \times \frac{W}{2^i}}, \\
x_f^i = relu(bn(conv(x_f^i))),\quad\quad x_f^i\in \mathbb{R}^{D_i\times \frac{H}{2^i} \times \frac{W}{2^i}}.
\end{gather}

\section{More Details of Experiments}

\subsection{More Details of Experiment Settings}

To customize the RIS methods~\cite{clipseg,mdetr} for the newly proposed RIM task, we made slight changes to the existing methods for a fair comparison. For CLIPSeg~\cite{clipseg}, we choose CLIP~\cite{clip} pre-trained \textbf{ViT-B/16}~\cite{vit} as the image encoder and set the projection dimension of the decoder as 64 (D=64). We add one sigmoid layer on the output to normalize it to standard matting output. For MDETR~\cite{mdetr}, we choose ResNet-101~\cite{he2016deep} as the image encoder, use the mask head as \textbf{smallconv}, and choose the prediction mask with the highest probability as the final output. For our CLIPMat, the channel numbers in matting semantic and details decoders are 768, 384, 192, and 96, respectively. We choose the CLIP~\cite{clip} pre-trained \textbf{ViT-B/16} and \textbf{ViT-L/14} as the image encoder, respectively. For \textbf{ViT-L/14}, we change the kernel size and stride of the patch embedding layer from 14 to 16 for a fair comparison. 

Both the CLIPSeg and MDETR use the weights that are further fine-tuned on VGPhraseCut~\cite{phrasecut}. However, CLIPMat only uses the CLIP pre-trained weights directly without further fine-tuning on VGPhraseCut~\cite{phrasecut} as we find it is unnecessary, which has also validated the value of our proposed RefMatte. For the parameters of position embedding that have a different shape from the pre-trained one, we reshape them by interpolation. The input size for all the methods is $512\times 512$, and we choose the largest batch size for each model, which is 32 for CLIPSeg, 8 for MDETR, 12 for CLIPMat(ViT/B-16), and 4 for CLIPMat(ViT/L-14). All the learning rates are fixed as $1e-4$. For CLIPSeg and MDETR, the image and text encoders are all frozen to follow the original design in their papers. For CLIPMat with \textbf{ViT/B-16}, the learning rates of the image encoder and the text encoder are $1e-6$ and $1e-7$, respectively. For CLIPMat with \textbf{ViT/L-14}, the learning rates of the image encoder and the text encoder are all $1e-6$. All the methods are trained 50 epochs on two NVIDIA A100 GPUs, which takes about 1 day for CLIPMat(ViT/B-16) and 3 days for CLIPMat(ViT/L-14). 

As for the optional matting refiner, we adopt the state-of-the-art automatic image matting model P3M from the work~\cite{p3m,p3mj}, modifying the input from a single image to the image with a coarse map. We train the refiner on RefMatte with the settings following the original paper to serve as an optional post-refiner in our case. We present both the visual results of CLIPMat with or without the refiner in the following section, showing that CLIPMat already performs very well without the matting refiner, although better results can be achieved with its help.

\subsection{Further Evaluation of the Main Results}
To provide a comprehensive evaluation of the results, besides the conventional evaluation metrics SAD (sum of absolute differences), MSE (mean squared error), and MAD (mean absolute difference), we also calculate the average SAD, MSE, and MAD for all the entities in each image and average them over the test set, denoted as \textbf{SAD(E)}, \textbf{MSE(E)}, and \textbf{MAD(E)}, respectively. Please note that \textit{SAD, MSE, MAD} are calculated on the basis of the foreground entities, which indicate the average error as per all foregrounds. However, \textit{SAD(E), MSE(E), MAD(E)} are calculated on the basis of the images, indicating the average error as per all images. Take the \textit{SAD} and \textit{SAD(E)} as examples, we show the details of them in Eq.~\eqref{eq:sad} and Eq.~\eqref{eq:sade}, where $N$ stands for the number of images and $M$ stands for the number of entities in one image (\eg, $M=2$ if the image contains a human and a dog). $G$ stands for the ground truth label and $P$ stands for the prediction. We simplify the SAD of each image as $|G-P|$. Thus, SAD(E), MSE(E), and MAD(E) reflect the models' ability to distinguish ambiguous foregrounds in the same image, serving as a more strict evaluation metric.
\begin{equation}
    SAD=\frac{1}{(N\times M)}\times\Sigma_1^n(\Sigma_1^m|G-P|),
    \label{eq:sad}
\end{equation}
\begin{equation}
    {SAD(E)}=\frac{1}{N}\times\Sigma_1^n(\frac{1}{M}\Sigma_1^m|G-P|).
    \label{eq:sade}
\end{equation}

\begin{figure*}[htbp]
    \centering
    \includegraphics[width=\linewidth]{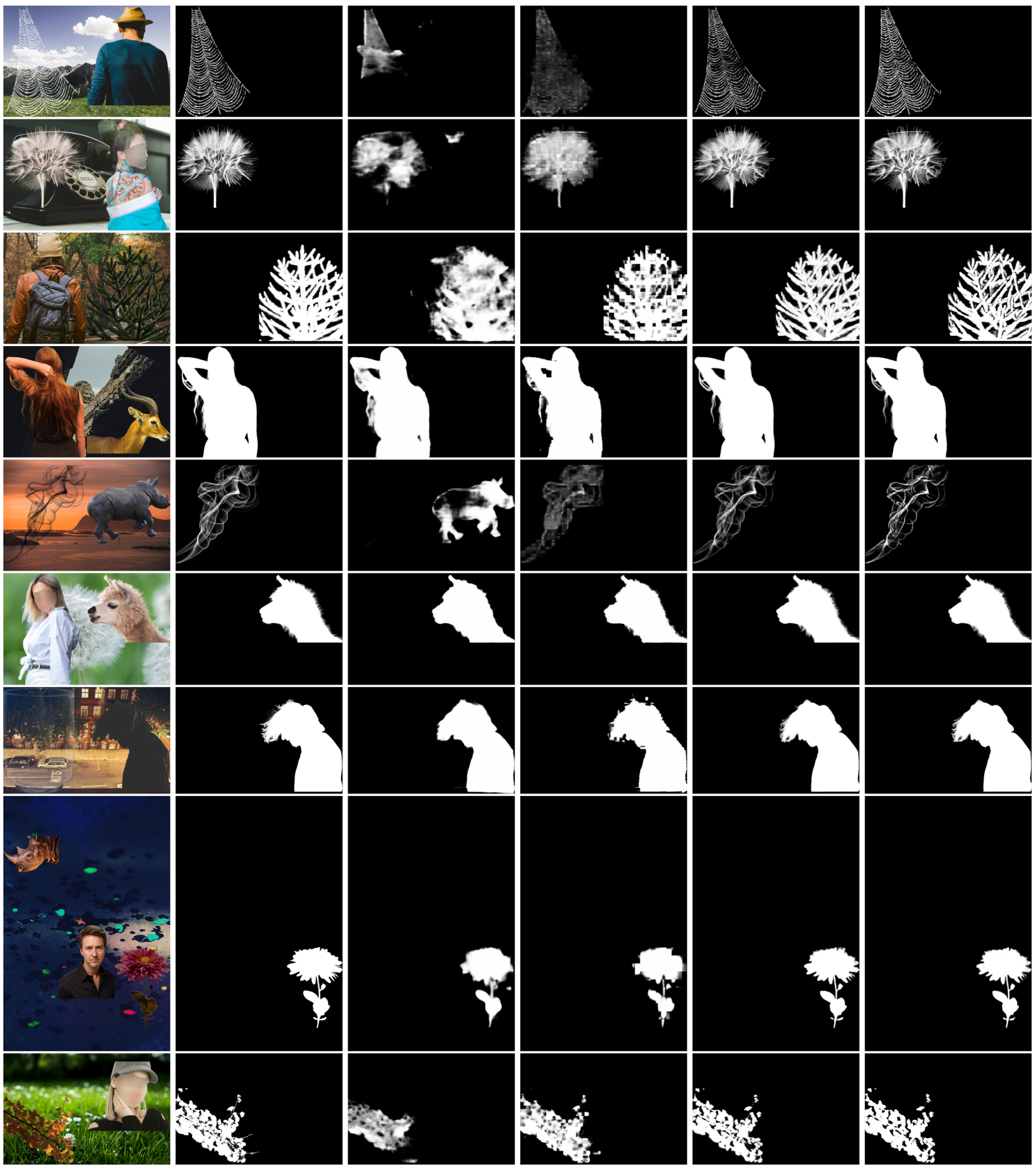}
    \caption{More subjective comparisons of different methods on the RefMatte under the keyword-based setting. From left to right: the original image, the ground truth, MDETR~\cite{mdetr}, CLIPSeg~\cite{clipseg}, our proposed CLIPMat, and CLIPMat with the matting refiner. The text inputs from the top to the bottom are: 1) net; 2) dandelion; 3) leaves; 4) human; 5) smog; 6) alpaca; 7) human; 8) flower; 9) leaves. We recommend zooming in for more details.}
    \label{fig:supp_keyword}
\end{figure*}

\begin{figure*}[htbp]
    \centering
    \includegraphics[width=0.92\linewidth]{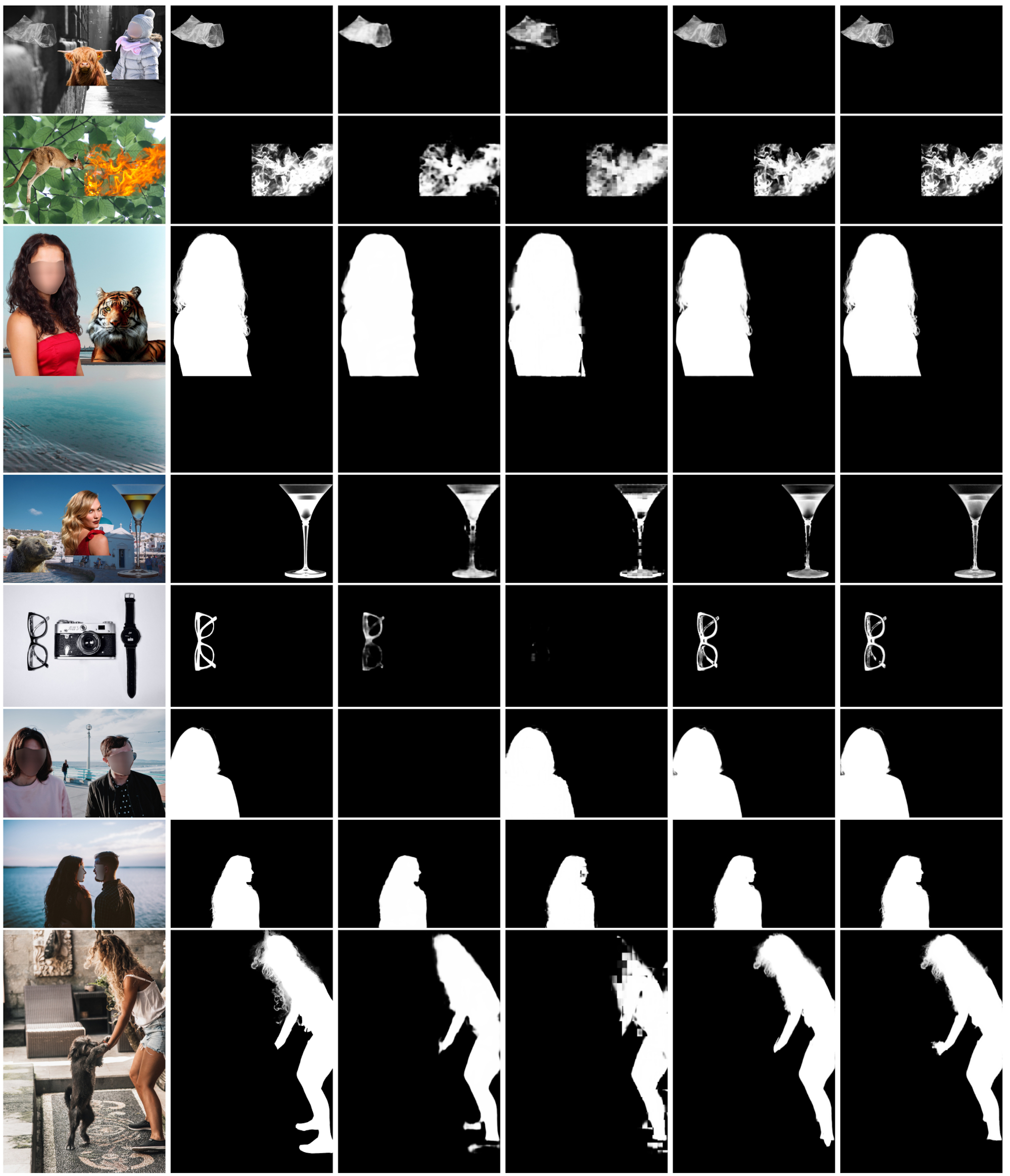}
    \caption{More subjective comparisons of different methods on the RefMatte under the expression-based setting and the RefMatte-RW100. From left to right: the original image, the ground truth, MDETR~\cite{mdetr}, CLIPSeg~\cite{clipseg}, our proposed CLIPMat, and CLIPMat with the matting refiner. The text inputs from the top to the bottom are: 1) the plastic bag which is lightgray and transparent; 2) the fire which is wheat at the most right side of the picture; 3) the woman in the crimson print on the left side of the rosybrown and salient brute; 4) the gray and salient vase at the most right side of the picture; 5) the good-looking glass on the left part of the photo; 6) a long-haired man in a pink shirt and sunglasses smiling; 7) a woman with long hair; 8) the lady in white clothes in the right of the picture. We recommend zooming in for more details.}
    \label{fig:supp_expression}
\end{figure*}

Here, we provide all the results of MDETR~\cite{mdetr}, CLIPSeg~\cite{clipseg}, and our proposed CLIPMat with or without the post-matting refiner in Table~\ref{tab:exp}. As can be seen, CLIPMat achieves the best results with both two backbones, where the larger backbone and the post-matting refiner improve the performance further. It is also noteworthy that the post matting refiner improves the results of MDETR and CLIPSeg by large margins, \ie, 32.27 to 27.33 for MDETR in keyword-based setting, 17.75 to 12.17 for CLIPSeg, but only improves a little bit for CLIPMat, \ie, 9.91 to 9.13 or 8.51 to 8.29. It owes to the excellent ability of CLIPMat to preserve details. Besides, for almost all the methods, SAD(E), MSE(E), and MAD(E) are larger than SAD, MSE, and MAD since they evaluate the ability of matting models to distinguish individual foreground in the same image, which is more challenging. However, CLIPMat's results on RefMatte-RW100 are even better in terms of SAD(E), MSE(E), and MAD(E), showing that CLIPMat has a good ability to extracting the correct targets. Furthermore, we provide more visual results to subjectively compare MDETR, CLIPSeg, and our proposed CLIPMat on the RefMatte test set and RefMatte-RW100 in both keyword and expression settings. The results are shown in the Figure~\ref{fig:supp_keyword} and Figure~\ref{fig:supp_expression}. As can be seen, CLIPMat performs very well in all the settings and outperforms all the other methods.

\begin{table*}[htbp]
\begin{center}
\small
\begin{tabular}{c|c|c|c|ccc|ccc}
\hline
 Setting & Method &Backbone &Refiner &SAD & MSE & MAD  & SAD(E) & MSE(E) & MAD(E) \\
 \hline
 \multirow{8}{*}{\makecell[c]{Keyword\\ setting}} & MDETR~\cite{mdetr} &  ResNet-101~\cite{he2016deep} & - & 32.27 &  0.0137& 0.0183 & 33.52& 0.0141&0.0190 \\
 \cline{2-10}
 & MDETR~\cite{mdetr} &  ResNet-101~\cite{he2016deep} & yes & 27.33 &  0.0123& 0.0155 & 28.22& 0.0126&0.0160 \\
 \cline{2-10}
 & CLIPSeg~\cite{clipseg} & ViT-B/16~\cite{vit} & - & 17.75& 0.0064&0.0101 & 18.69& 0.0067&0.0106 \\
 \cline{2-10}
 & CLIPSeg~\cite{clipseg} & ViT-B/16~\cite{vit} & yes & 12.17 &0.0042 &0.0069 & 12.75& 0.0044&0.0073\\
 \cline{2-10}
 & CLIPMat & ViT-B/16 & -  & 9.91 & 0.0028 & 0.0057  &10.41 &0.0029 &0.0059 \\
 \cline{2-10}
 & CLIPMat & ViT-B/16 & yes  & 9.13 & 0.0026 & 0.0052 & 9.56&0.0027 &0.0055 \\
 \cline{2-10}
 & CLIPMat & ViT-L/14 & -  & 8.51 & 0.0022 & 0.0049  & 8.98&0.0023 &0.0051 \\
\cline{2-10}
& CLIPMat & ViT-L/14 & yes  & \textbf{8.29} & \textbf{0.0022} & \textbf{0.0027}  & \textbf{8.72}& \textbf{0.0023}& \textbf{0.0050}\\
\hline
 \multirow{8}{*}{\makecell[c]{Expression \\ setting}} & MDETR~\cite{mdetr} &  ResNet-101~\cite{he2016deep} & - & 84.70 &0.0434 &0.0482 &90.45 &0.0463 &0.0515\\
 \cline{2-10}
 & MDETR~\cite{mdetr} &  ResNet-101~\cite{he2016deep} & yes & 80.48 &  0.0424& 0.0458 & 85.83& 0.0452&0.0488 \\
 \cline{2-10}
 & CLIPSeg~\cite{clipseg} & ViT-B/16~\cite{vit} & - & 69.13& 0.0358& 0.0394 & 73.53 & 0.0381& 0.0419\\
 \cline{2-10}
 & CLIPSeg~\cite{clipseg} & ViT-B/16~\cite{vit} & yes & 64.48&0.0341 &0.0367 & 68.56& 0.0364&0.0391\\
 \cline{2-10}
 & CLIPMat & ViT-B/16 & -  & 47.97 & 0.0245 & 0.0273  &50.84 &0.0260 &0.0273 \\
 \cline{2-10}
 & CLIPMat & ViT-B/16 & yes  & 46.38 & 0.0239 & 0.0264 &49.11 &0.0253 &0.0279 \\
 \cline{2-10}
 & CLIPMat & ViT-L/14 & -  & 42.05 & 0.0212 & 0.0238   & 44.77& 0.0226& 0.0254\\
\cline{2-10}
& CLIPMat & ViT-L/14 & yes  & \textbf{40.37} & \textbf{0.0205} & \textbf{0.0229}  & \textbf{43.03}& \textbf{0.0218}& \textbf{0.0244}\\

\hline
\multirow{8}{*}{\makecell[c]{RefMatte- \\ RW100}} & MDETR~\cite{mdetr} &  ResNet-101~\cite{he2016deep} & - & 131.58 &0.0675 &0.0751 & 136.59& 0.0700&0.0779 \\
 \cline{2-10}
 & MDETR~\cite{mdetr} & 
 ResNet-101~\cite{he2016deep} & yes & 125.78 &  0.0669& 0.0717 & 130.72& 0.0697&0.0744 \\
 \cline{2-10}
 & CLIPSeg~\cite{clipseg} & ViT-B/16~\cite{vit} & - & 211.86 &0.1178&0.1222 &222.37 &0.1236 &0.1282 \\
  \cline{2-10}
 & CLIPSeg~\cite{clipseg} & ViT-B/16~\cite{vit} & yes & 207.04&0.1166 &0.1195 & 216.93& 0.1222&0.1252\\
 \cline{2-10}
 & CLIPMat & ViT-B/16 & -  & 110.66 & 0.0614 & 0.0636  &110.63 &0.0612 & 0.0635  \\
 \cline{2-10}
 & CLIPMat & ViT-B/16 & yes  & 107.81 & 0.0595 & 0.0620 & 107.23& 0.0591& 0.0616\\
 \cline{2-10}
 & CLIPMat & ViT-L/14 & -  & 88.52 & 0.0488 & 0.0510 & 87.92& 0.0483& 0.0505\\
\cline{2-10}
& CLIPMat & ViT-L/14 & yes  & \textbf{85.83} & \textbf{0.0474} & \textbf{0.0495} & \textbf{84.93}& \textbf{0.0468}& \textbf{0.0488}\\
\hline
\end{tabular}
\caption{Results on the RefMatte test set in two settings and the RefMatte-RW100 test set, a.k.a the complete version of Table 2 in the paper.}
\label{tab:exp}
\end{center}
\end{table*}

\subsection{More Ablation Studies}

\subsubsection{Error Bars from Multiple Runs with Different Seeds} 
To calculate the error bars of our proposed method CLIPMat on RefMatte, we run the experiments on different random seeds and test them on both the keyword-based and expression-based settings of the RefMatte test set and RefMatte-RW100. We then report all the results in Table~\ref{tab:error_bars} by calculating the Standard Deviation (Std.) and mean value of the SAD. As can be seen, CLIPMat performs very stably on both the RefMatte test set and RefMatte-RW100.

\begin{table}[htbp]
\begin{center}
\begin{tabular}{c|c|cccccc}
\hline
Dataset & Setting & SAD & MSE & MAD & \multicolumn{1}{|c}{SAD-E} & MSE-E & MAD-E \\
\hline
 \multirow{4}{*}{\makecell[c]{RefMatte\\test set}} & \multirow{4}{*}{keyword}& \underline{9.91}& 0.0028& 0.0057& \multicolumn{1}{|c}{10.41}& 0.0029&0.0059\\
 & &\underline{9.82} & 0.0028 & 0.0056 & \multicolumn{1}{|c}{10.40} & 0.0029 & 0.0059 \\
 & &\underline{10.06}& 0.0030& 0.0057& \multicolumn{1}{|c}{10.71}&0.0031 & 0.0061\\
 \cline{3-8}
 && \textit{Error bar of SAD} & \multicolumn{1}{c}{\textit{Mean:}} & \textit{9.93} &  \textit{Std.:} &\textit{0.1212} & \\
 \hline
 \multirow{4}{*}{\makecell[c]{RefMatte\\test set}}&\multirow{4}{*}{expression}&\underline{47.97} &0.0245 &0.0273 &\multicolumn{1}{|c}{50.84} & 0.0260& 0.0273\\
 & &\underline{42.52}&0.0215 &0.0241 &\multicolumn{1}{|c}{45.50} &0.0231 & 0.0258\\
 & &\underline{50.35}& 0.0259&0.0287 &\multicolumn{1}{|c}{53.09} &0.0273 &0.0303 \\
 \cline{3-8}
 && \textit{Error bar of SAD} & \multicolumn{1}{c}{\textit{Mean}} & \textit{46.95} &  \textit{Std.:} &\textit{4.0141} & \\
 \hline
 \multirow{4}{*}{\makecell[c]{RefMatte\\-RW100}}&\multirow{4}{*}{expression}& \underline{110.66} &0.0614 &0.0636 &\multicolumn{1}{|c}{110.63} &0.0612& 0.0635 \\
 & &\underline{121.21}&0.0676 &0.0698 &\multicolumn{1}{|c}{119.65} &0.0667 & 0.0690\\
 & &\underline{117.79}& 0.0657& 0.0677&\multicolumn{1}{|c}{120.46} & 0.0670&0.0691 \\
 \cline{3-8}
 && \textit{Error bar of SAD} & \multicolumn{1}{c}{\textit{Mean:}} & \textit{116.55} &  \textit{Std.:} &\textit{5.3826} & \\
 \hline
\end{tabular}
\end{center}
\caption{Results of CLIPMat with different random seeds on the RefMatte test set and RefMatte-RW100.}
\label{tab:error_bars}
\end{table}

\subsubsection{Impact of Input Texts}

\textbf{Impact of prompt templates} We trained CLIPMat with our proposed matting-related pre-embedding context to enhance the robustness of different input prompt templates. Here We investigate the impact of different prompt templates of CLIPMat(ViT-B/16) and show the results in Table~\ref{tab:ablation_prompt}. The default setting for our previous experiments in the main results is \textit{a photo of a $\left\{ keyword \right\}$}. As can be seen, CLIPMat is robust to all kinds of prompt templates while achieving the best results in the setting \textit{an image of a $\left\{ keyword \right\}$}, and also performs very well on matting-related prompt templates, \eg, \textit{the foreground of the $\left\{ keyword \right\}$}, \textit{to extract $\left\{ keyword \right\}$} and so on. We believe the success is owing to our careful design of matting-related pre-context embedded prompts. More efforts on prompt augmentation could be made in future work to further improve the performance.

\begin{table*}[htbp]
\begin{center}
\begin{tabular}{c|ccc|ccc}
\hline
prompt template & SAD & MSE & MAD & SAD(E) & MSE(E) & MAD(E) \\
\hline
\textit{$\left\{ keyword \right\}$} & 9.95& 0.0029& 0.0057& 10.45& 0.0030& 0.0060\\
\textit{a photo of a $\left\{ keyword \right\}$} & 9.91 &0.0028 &0.0057&10.41 &0.0029&0.0059 \\
\textit{a photograph of a $\left\{ keyword \right\}$} &9.82 & 0.0028& 0.0056& 10.32&0.0029&0.0059\\
\textit{an image of a $\left\{ keyword \right\}$} &9.84 &0.0028 &0.0056 &10.34 &0.0029 &0.0059 \\
\textit{the foreground of the $\left\{ keyword \right\}$} & 9.87& 0.0028& 0.0056& 10.36&0.0029 &0.0059 \\
\textit{the mask of the $\left\{ keyword \right\}$} &9.90&0.0028 &0.0057 &10.39 &0.0029 &0.0059 \\
\textit{the alpha matte of the $\left\{ keyword \right\}$} & 10.01& 0.0029& 0.0057&10.50 &0.0030 &0.0060 \\
\textit{to extract $\left\{ keyword \right\}$} &9.88&0.0028& 0.0056& 10.35&0.0029 &0.0059 \\
\hline
\end{tabular}
\caption{Results of CLIPMat with different prompt templates on the RefMatte test set in the keyword-based setting.}
\label{tab:ablation_prompt}
\end{center}
\end{table*}

\noindent\textbf{Impact of different expressions} Since we have introduced different types of expressions in our task, it is interesting to investigate the influence of each type on matting performance. As shown in Table~\ref{tab:ablation_expression}, we evaluate CLIPMat(ViT-B/16) on the RefMatte test set and RefMatte-RW100. As can be seen, the relative position expression is more informative (or easy to understand) than others, leading to the best performance on both the synthetic test set as well as the real-world one. Among them, \textit{RPE-1}, which is shorter and more straightforward compared with \textit{RPE-2} achieves the best result in the expression-based setting. On the other hand, the absolute position expression has worse performance compared with relative ones in the expression-based setting but has comparable performance in RefMatte-RW100, probably due to manually labeled annotations being more straightforward and meaningful. Another interesting finding is that CLIPMat performs worse on basic expression than position-based expression, which is counter-intuitive. We believe the reason is that the model has been trained towards emphasizing the relationship between entities, resulting in a relatively poor ability to understand the entity and its own attributes. For the RefMatte-RW100 dataset, \textit{FREE} prompts, which are labeled by human annotators following their preferred style, lead to the worst results, mainly due to the significant diversity of logic, grammar, and words used to describe the entities. More efforts could be made to study the most effective expressions that matter in automatic applications as well as improve the generalization ability of RIM models to deal with diverse expressions in human-machine interaction applications.

\begin{table*}[htbp]
\begin{center}
\begin{tabular}{c|c|ccc|ccc}
\hline
Setting&prompt template & SAD & MSE & MAD & SAD(E) & MSE(E) & MAD(E) \\
\hline
\multirow{4}{*}{\makecell[c]{Expression-based \\ setting}}&\textit{BE} & 59.20& 0.0308& 0.0337& 62.81& 0.0327& 0.0357\\
&\textit{APE} &59.41 &0.0310 &0.0339 & 62.81&0.0328 &0.0358 \\
&\textit{RPE-1} &25.98 &0.0120 &0.0147 &28.14 &0.0131 &0.0159 \\
&\textit{RPE-2} & 46.29& 0.0236& 0.0264& 48.33& 0.0246& 0.0275\\
\hline
\multirow{4}{*}{\makecell[c]{RefMatte-RW100 \\ dataset}}&\textit{BE} & 152.11& 0.0857&0.0880 &156.44 &0.0881 &0.0905 \\
&\textit{APE} & 86.01& 0.0469& 0.0490& 84.49&0.0458 &0.0480 \\
&\textit{RPE} &85.08 & 0.0458& 0.0479& 83.59& 0.0446& 0.0469\\
&\textit{FREE} & 157.34& 0.0883& 0.0905& 161.29& 0.0905& 0.0928\\
\hline
\end{tabular}
\caption{Expression-based RIM results on RefMatte and RefMatte-RW100. BE: basic expression. APE: absolute position expression. RPE: relative position expression. FE: free expression.}
\label{tab:ablation_expression}
\end{center}
\end{table*}

\subsubsection{The Impact of Pre-training and Freezing}

We further investigate the impact of pre-training models and freezing the parameters by conducting the ablation studies on the keyword-based setting of RefMatte with CLIPMat(ViT-B/16) and presenting the results in Table~\ref{tab:pretrain}. As can be seen, freezing the parameters of CLIP~\cite{clip} pre-trained image encoder and text encoder results in SAD of 20.05, while fine-tuning the parameters as our proposed method results in SAD of 9.91. We hypothesize the benefits come from reducing the visual and text gap between the RefMatte and the CLIP dataset through fine-tuning. We set the learning rate of the image encoder as 0.01 times of matting decoders, and the learning rate of the text encoder as 0.001 times of matting decoders. On the other hand, pre-training CLIPMat on VGPhraseCut~\cite{phrasecut} does not provide performance improvement for CLIPMat, \ie, SAD of 11.33 v.s. SAD of 9.91. Such results have also confirmed the value of RefMatte since training on it directly can achieve even better results, implying that the proposed CLIPMat can serve as a simple and strong baseline for referring image matting.

\begin{table}[htbp]
\begin{center}
\begin{tabular}{ccc|ccc|ccc}
\hline
CLIP-Pretrain & CLIP-Freeze & Phrasecut-Pretrain  & SAD & MSE & MAD & SAD(E) & MSE(E) & MAD(E) \\
\hline
\checkmark& \checkmark& & 20.05 & 0.0084 & 0.0115 &21.39&0.0089 & 0.0122\\
& & \checkmark& 11.33 & 0.0036 & 0.0065 &11.86&0.0037 & 0.0068\\
\checkmark& & & 9.91 & 0.0028 & 0.0057 &10.41&0.0029 & 0.0059\\
 \hline
\end{tabular}
\end{center}
\caption{Ablation studies of freezing the pre-trained CLIP~\cite{clip} parameters and using VGPhraseCut~\cite{phrasecut} pre-trained weights.}
\label{tab:pretrain}
\end{table}

\subsection{Failure Cases}

\begin{figure}[htbp]
    \centering
    \includegraphics[width=1\linewidth]{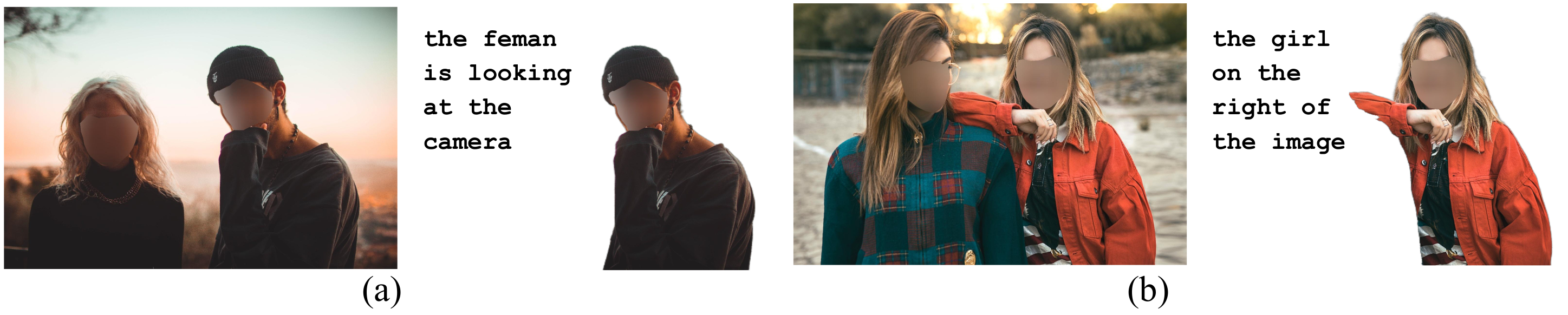}
    \caption{Some failure cases of CLIPMat, which is trained on RefMatte and tested on RefMatte-RW100. (a) Incorrect foreground instance. (b) Incomplete foreground details.}
    \label{fig:failure_case}
\end{figure}

Although our proposed CLIPMat shows good performance on both the synthetic images and the real-world images after training on the RefMatte dataset, it still encounters some failure cases. We present some failure cases in Figure~\ref{fig:failure_case} (a) and (b). As shown in the figure, CLIPMat fails to locate the accurate foreground under some complex or ambiguous expression guidance. For example, as in (a), CLIPMat may not understand the word \textit{feman} correctly and focus more on \textit{looking at the camera}. In some other cases, like shown in (b) in the figure, CLIPMat fails to extract all the details of the foreground, which indicates the ability to preserve local details can be further improved. Such failure cases can be improved by 1) enhancing CLIPMat's abilities in understanding complex expressions and segmenting the foregrounds out with detailed boundaries, especially for those which have occlusions with other entities; 2) reducing the domain gap between synthetic and real-world images/expressions. We leave them as future work.

{\small
\bibliographystyle{ieee_fullname}
\bibliography{supp}
}

\newpage

\begin{figure*}[htbp]
    \centering
    \includegraphics[width=0.95\linewidth]{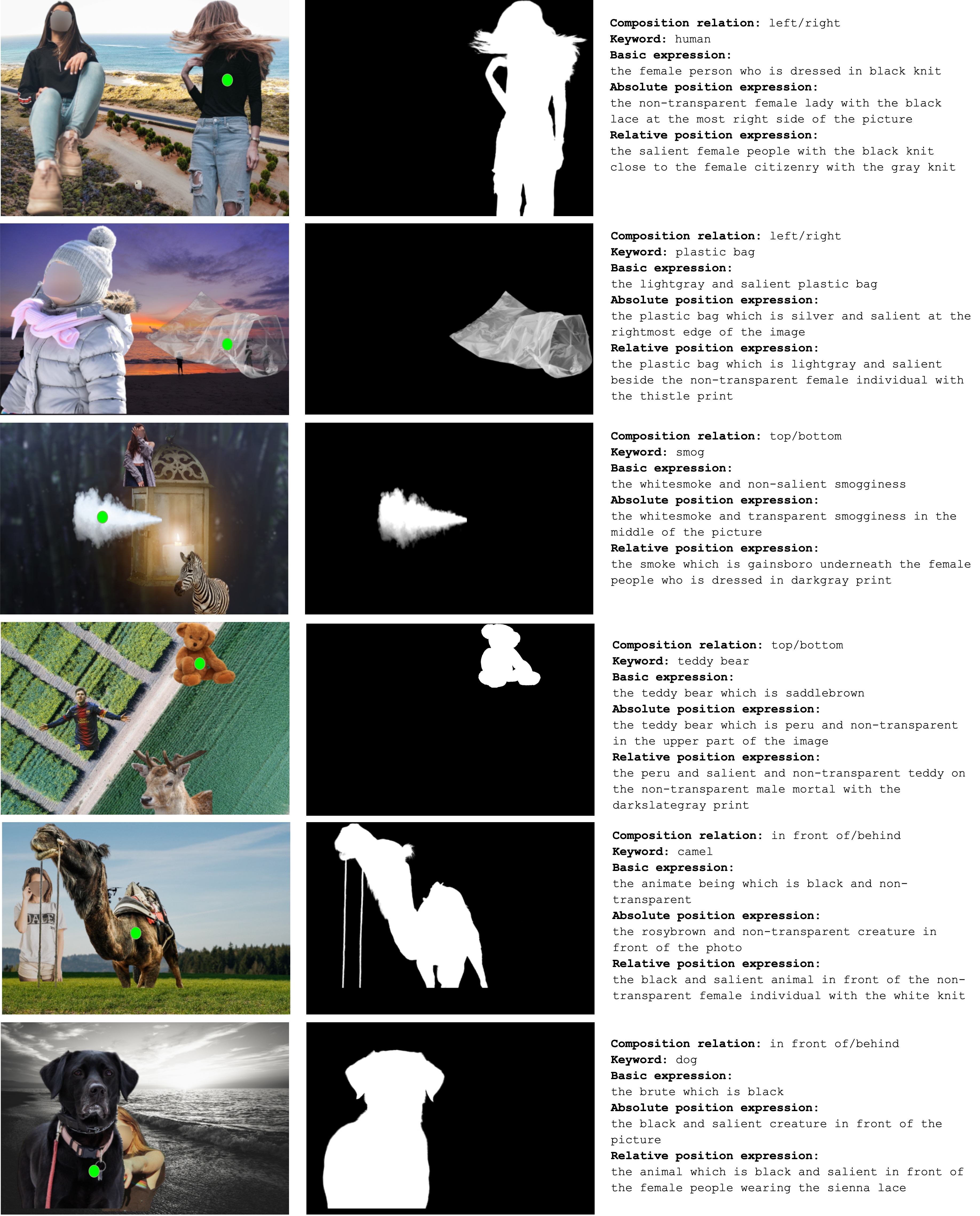}
    \caption{More examples from our RefMatte dataset. The first column shows the composite images with different foreground instances, and the second column and the third column show the ground truth alpha mattes and the natural language descriptions corresponding to the specific instances indicated by the green dots, respectively.}
    \label{fig:refmatte_supp}
\end{figure*}

\begin{figure}[htbp]
    \centering
    \includegraphics[width=0.97\linewidth]{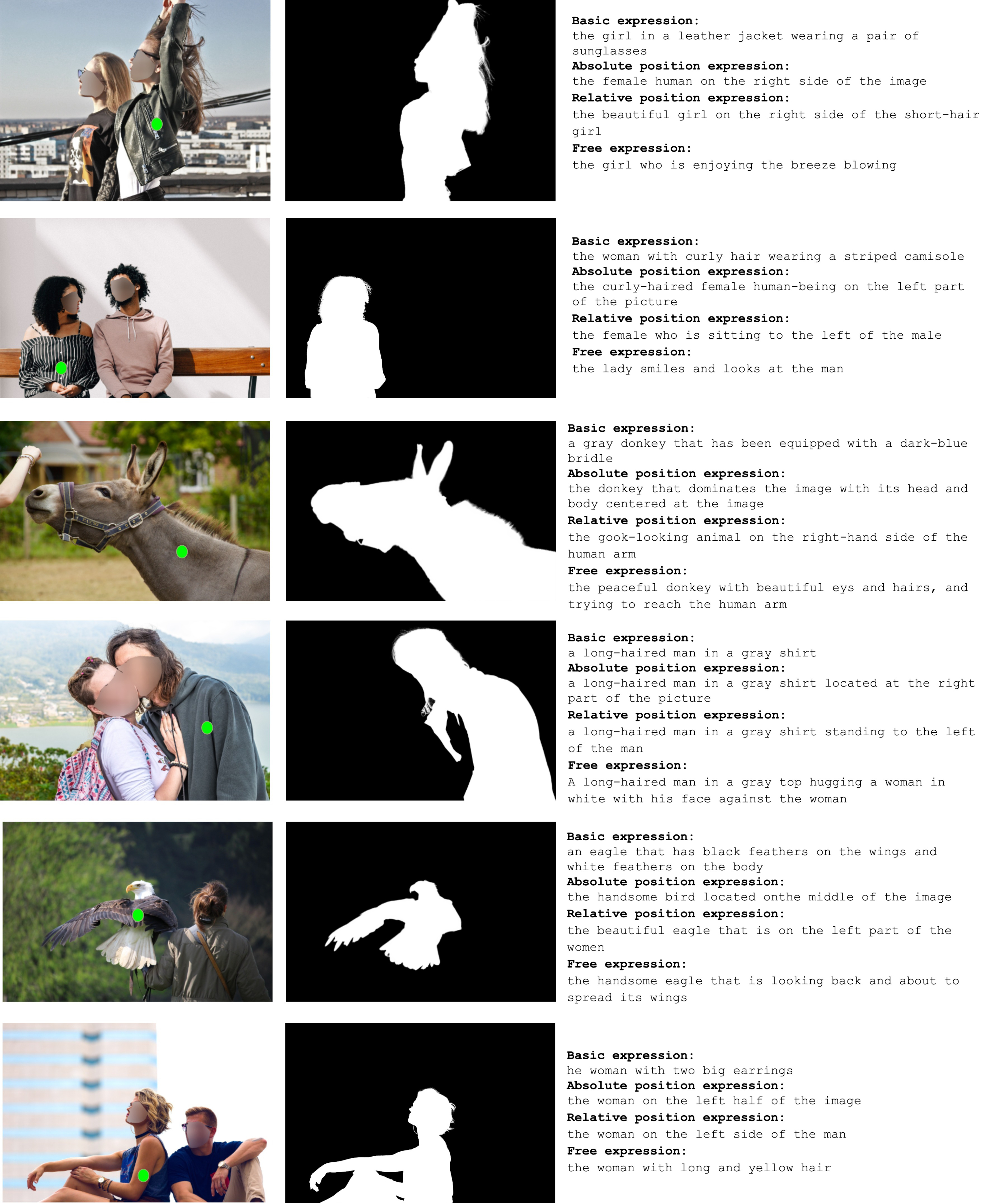}
    \caption{More examples from our RefMatte-RW100. The first column shows real-world images with different foreground instances, and the second column and the third column show the ground truth alpha mattes and the natural language descriptions corresponding to the specific instances indicated by the green dots, respectively.}
    \label{fig:refmatte_rw}
\end{figure}

\begin{table*}
\begin{center}
\small
\begin{tabular}{|c|c|c|}
\hline
Block name & Output size & Detail  \\
\hline
\multicolumn{3}{|c|}{\textbf{CLIP - Text Encoder}}  \\
\hline
$embed$ & $N\times22 (77)\times512$ & concatenate (text, context) + position embedding \\
\hline
$T_1$ &  $N \times 22 (77)\times512$ & transformer block (heads 8, width 512)\\
\hline
\multicolumn{3}{|c|}{\textbf{CLIP - Image Encoder}}  \\
\hline
$conv1$ & $N\times768\times32\times32$ & conv ($3\times3$, 768, stride 16) + BN + ReLU\\
\hline
$embed$ & $1025 \times N\times768$ & + class embedding + position embeding \\
\hline
$T_1$ & $1025 \times N\times768$ & transformer block (heads 12, width 768) $\times$ 3\\
\hline
$T_2$ & $1025 \times N\times768$ & transformer block (heads 12, width 768) $\times$ 3\\
\hline
$T_3$ & $1025 \times N\times768$ & transformer block (heads 12, width 768) $\times$ 3\\
\hline
$T_4$ & $1025 \times N\times768$ & transformer block (heads 12, width 768) $\times$ 3\\
\hline
\multicolumn{3}{|c|}{\textbf{TSP}}\\
\hline
$norm\_visual$ & $N\times1025\times256$ & LN + Linear(256) + LN\\
\hline
$norm\_text$ & $N\times22 (77)\times256$ & LN + Linear(256) + LN\\
\hline
$cross\_attn$ & $N\times1025\times256$ & attention (256, heads=4, dropout=0.1)\\
\hline
$self\_attn$ & $N\times1025\times256$ & attention (256, heads=4, dropout=0.1)\\
\hline
$out$ & $N\times64\times32\times32$ & conv ($1\times1$, 64, stride 1)\\
\hline
\multicolumn{3}{|c|}{\textbf{Matting Semantic Decoder}}  \\
\hline
$D_2$ & $N\times32\times128\times128$ & \tabincell{c}{[ conv ($3\times3$, 32, stride 3) + BN + ReLU ] $\times$ 2\\ upsample(4)}\\
\hline
$D_1$ & $N\times32\times512\times512$ & \tabincell{c}{[ conv ($3\times3$, 32, stride 3) + BN + ReLU ] $\times$ 2\\ upsample(4)}\\
\hline
$D_0$ & $N\times3\times512\times512$ & conv ($3\times3$, 3, stride 1) \\
\hline
\multicolumn{3}{|c|}{\textbf{MDE}}\\
\hline
$norm\_visual$ & $N\times384\times \frac{H}{2^i} \times \frac{H}{2^i}$ & \tabincell{c}{conv ($1\times1$, 384, stride 1)\\ upsample($2^{4-i}$)} \\
\hline
$norm\_img$ & $N\times384\times \frac{H}{2^i} \times \frac{H}{2^i}$ & \tabincell{c}{conv ($1\times1$, 384, stride 1)\\ maxpool ($2^i$)} \\
\hline
$out$ & $N\times{D_i}\times \frac{H}{2^i} \times \frac{H}{2^i}$ & \tabincell{c}{concatenate (visual, img)\\conv ($3\times3$, $D_i$, stride 1) + BN + ReLU\\} \\
\hline
\multicolumn{3}{|c|}{\textbf{Matting Details Decoder}}  \\
\hline
$D_4$ & $N\times384\times64\times64$ & \tabincell{c}{[ conv ($3\times3$, 384, stride 3) + BN + ReLU ] $\times$ 2\\ upsample(2)}\\
\hline
$D_3$ & $N\times192\times128\times128$ & \tabincell{c}{[ conv ($3\times3$, 192, stride 3) + BN + ReLU ] $\times$ 2\\ upsample(2)}\\
\hline
$D_2$ & $N\times96\times256\times256$ & \tabincell{c}{[ conv ($3\times3$, 96, stride 3) + BN + ReLU ] $\times$ 2\\ upsample(2)}\\
\hline
$D_1$ & $N\times32\times512\times512$ & \tabincell{c}{[ conv ($3\times3$, 32, stride 3) + BN + ReLU ] $\times$ 2\\ upsample(2)}\\
\hline
$D_0$ & $N\times1\times512\times512$ & conv ($3\times3$, 1, stride 1) \\

\hline
\multicolumn{3}{|c|}{\textbf{Collaborative Matting}}  \\
\hline
$CM$ & $N\times1\times512\times512$ & pixel-wise multiply for output from two matting decoders output .\\
\hline

\end{tabular}
\caption{Network structure of our proposed CLIPMat, where $N$ stands for batch size. The input of CLIPMat is a batch of images of the size $N\times3\times512\times512$, and a batch of text descriptions of size $N\times14$ for keyword-setting and $N\times69$ for expression-setting as well as the learnable context with size $N\times 8 \times 512$.}
\label{Tab:clipmat_structure}
\end{center}
\end{table*}

\clearpage
\newpage
\appendix

\section{Datasheet of RefMatte}

\subsection{Motivation}

\noindent \textbf{1. For what purpose was the dataset created? Was there a specific task in mind? Was there a specific gap that needed to be filled? Please provide a description.}

\textbf{A1:} RefMatte is created to facilitate the study of a new task: referring image matting (RIM). RIM is first introduced in this paper as extracting the meticulous foreground in the image with linguistic keyword or expression as an auxiliary input. However, prevalent visual grounding methods are all limited to the segmentation level, probably due to the lack of high-quality datasets for RIM. To fill the gap, we establish the first large-scale challenging dataset \textbf{RefMatte} by designing a comprehensive image composition and expression generation engine to produce synthetic images on top of current public high-quality matting foregrounds with flexible logics and re-labelled diverse attributes. 

\noindent \textbf{2. Who created this dataset (e.g., which team, research group) and on behalf of which entity (e.g., company, institution, organization)?}

\textbf{A2:} RefMatte is created by the authors.

\noindent \textbf{3. Who funded the creation of the dataset? If there is an associated grant, please provide the name of the grantor and the grant name and number.}

\textbf{A3:} This study was supported by Australian Research Council Projects in part by FL170100117 and IH180100002.

\subsection{Composition}

\noindent \textbf{1. What do the instances that comprise the dataset represent (e.g., documents, photos, people, countries)? Are there multiple types of instances(e.g., movies, users, and ratings; people and interactions between them; nodes and edges)? Please provide a description.}

\textbf{A1:} The RefMatte dataset consists of images covering 230 categories which are very popular in the field of image matting, and expressions describe each entity. Some typical types are humans, animals, plants, spiders web and so on. All the personally identifiable information has been preserved for privacy consent.

\noindent \textbf{2. How many instances are there in total (of each type, if appropriate)?}

\textbf{A2:} The RefMatte dataset contains 230 object categories, 47,500 images, 118,749 expression-region entities with high-quality alpha matte, and 474,996 expressions.

\noindent \textbf{3. Does the dataset contain all possible instances or is it a sample (not necessarily random) of instances from a larger set? If the dataset is a sample, then what is the larger set? Is the sample representative of the larger set (e.g., geographic coverage)? If so, please describe how this representativeness was validated/verified. If it is not representative of the larger set, please describe why not (e.g., to cover a more diverse range of instances, because instances were withheld or unavailable).}

\textbf{A3:} The RefMatte itself is a large set that contains a large number of instances. It is large enough to be used for training deep models. However, the composition and expression engines we designed make it possible and easy to extend the dataset to a larger scale.

\noindent \textbf{4. What data does each instance consist of? “Raw” data (e.g., unprocessed text or images)or features? In either case, please provide a description.}

\textbf{A4:} Each instance consists of a high-resolution synthetic image generated by our composition engine, the high-quality alpha matte of the specific entity, the keyword and the expressions that used to describe the specific entity.

\noindent \textbf{5. Is there a label or target associated with each instance? If so, please provide a description.}

\textbf{A5:} Yes. Each instance is associated with a label, including an alpha matte, a keyword name, and several expressions. Some examples can be seen from the Figure~\ref{fig:refmatte_supp} and Figure~\ref{fig:refmatte_rw}.

\noindent \textbf{6. Is any information missing from individual instances? If so, please provide a description, explaining why this information is missing (e.g., because it was unavailable). This does not include intentionally removed information, but might include, e.g., redacted text.}

\textbf{A6:} No.

\noindent \textbf{7. Are relationships between individual instances made explicit (e.g., users’ movie ratings, social network links)? If so, please describe how these relationships are made explicit.}

\textbf{A7:} Yes. We keep a JSON file to store the relationships between individual instances, i.e., information such as whether or not multiple instances are located on the same image.

\noindent \textbf{8. Are there recommended data splits (e.g., training, development/validation, testing)? If so, please provide a description of these splits, explaining the rationale behind them.}

\textbf{A8:} Yes. We split the dataset RefMatte into training and test sets manually. We keep all the long-tail categories in the training set only. More details about the data splits can be found in Section 3.3 of the paper.

\noindent \textbf{9. Are there any errors, sources of noise, or redundancies in the dataset? If so, please provide a description.}

\textbf{A9:} Although we have manually checked the annotation information very carefully, there may be some minor inaccurate expression labels. However, since linguistic expression generated by real human is also subjective and may contains some error. We believe this might be a source of noise to improve the generalization ability of trained models.

\noindent \textbf{10. Is the dataset self-contained, or does it link to or otherwise rely on external resources (e.g., websites, tweets, other datasets)? If it links to or relies on external resources, a) are there guarantees that they will exist, and remain constant, over time; b) are there official archival versions of the complete dataset (i.e., including the external resources as they existed at the time the dataset was created); c) are there any restrictions (e.g., licenses, fees) associated with any of the external resources that might apply to a future user? Please provide descriptions of all external resources and any restrictions associated with them, as well as links or other access points, as appropriate.}

\textbf{A10:} The RefMatte dataset is comprised of the publicly available datasets, including AM-2k~\cite{gfm}, P3M-10k~\cite{p3m}, AIM-500~\cite{aim}, SIM~\cite{sim}, DIM~\cite{dim}, and HATT~\cite{hatt}. These datasets are publicly available and can be downloaded from their websites. We appreciate the significant contribution of the authors to the research community. We show the details of each matting dataset in Section~\ref{sec:matting_datasets}. 

\noindent \textbf{11. Does the dataset contain data that might be considered confidential (e.g., data that is protected by legal privilege or by doctorpatient confidentiality, data that includes the content of individuals non-public communications)? If so, please provide a description.}

\textbf{A11:} No.

\noindent \textbf{12. Does the dataset contain data that, if viewed directly, might be offensive, insulting, threatening, or might otherwise cause anxiety? If so, please describe why.}

\textbf{A12:} No.

\subsection{Collection Process}

\noindent \textbf{1. How was the data associated with each instance acquired? Was the data directly observable (e.g., raw text, movie ratings), reported by subjects (e.g., survey responses), or indirectly inferred/derived from other data (e.g., part-of-speech tags, model-based guesses for age or language)? If data was reported by subjects or indirectly inferred/derived from other data, was the data validated/verified? If so, please describe how.}

\textbf{A1:} The data associated with each instance are generated through a semi-automatic style by combining the attribute information predicted by pretrained models and manual annotations. We report the details in Section 3.1 of the paper.

\noindent \textbf{2. What mechanisms or procedures were used to collect the data (e.g., hardware apparatus or sensor, manual human curation, software program, software API)? How were these mechanisms or procedures validated?}

\textbf{A2:} The matting entities in the dataset RefMatte are collected from publicly available datasets described above, which can be directly downloaded from their websites. The final images are generated by our own proposed composition engine with the details described in Section 3.2 of the paper.

\noindent \textbf{3. If the dataset is a sample from a larger set, what was the sampling strategy (e.g., deterministic, probabilistic with specific sampling probabilities)?}

\textbf{A3:} No.

\noindent \textbf{4. Who was involved in the data collection process (e.g., students, crowdworkers, contractors) and how were they compensated (e.g., how much were crowdworkers paid)?}

\textbf{A4:} The first author of this paper.

\noindent \textbf{5. Over what timeframe was the data collected? Does this timeframe match the creation timeframe of the data associated with the instances (e.g., recent crawl of old news articles)? If not, please describe the timeframe in which the data associated with the instances was created.}

\textbf{A5}: It took about 30 days to collect the data and about 2 months to complete organization and annotation.

\subsection{Preprocessing/cleaning/labeling}

\noindent \textbf{1. Was any preprocessing/cleaning/labeling of the data done (e.g., discretization or bucketing, tokenization, part-of-speech tagging, SIFT feature extraction, removal of instances, processing of missing values)? If so, please provide a description. If not, you may skip the remainder of the questions in this section.}

\textbf{A1:} Yes. First, we collect and clean the data from currently available matting datasets to serve as matting entities, then we use our proposed composition and expression engine to generate the synthetic images with linguistic labels. The details can be seen in Section 3 of the paper.

\noindent \textbf{2. Was the ``raw" data saved in addition to the preprocessed/cleaned/labeled data (e.g., to support unanticipated future uses)? If so, please provide a link or other access point to the ``raw" data.}

\textbf{A2:} N/A.

\noindent \textbf{3. Is the software used to preprocess/clean/label the instances available? If so, please provide a link or other access point.}

\textbf{A3:} No. We process all the data with our code, which will be released.

\subsection{Uses}

\noindent \textbf{1. Has the dataset been used for any tasks already? If so, please provide a description.}

\textbf{A1:} No.

\noindent \textbf{2. Is there a repository that links to any or all papers or systems that use the dataset? If so, please provide a link or other access point.}

\textbf{A2:} N/A.

\noindent \textbf{3. What (other) tasks could the dataset be used for?}

\textbf{A3:} The RefMatte dataset can be used for referring image matting studies. In addition, it can be used for machine learning topics like domain adaptation, referring image localization, and one-shot/zero-shot referring image matting.

\noindent \textbf{4. Is there anything about the composition of the dataset or the way it was collected and preprocessed/cleaned/labeled that might impact future uses? For example, is there anything that a future user might need to know to avoid uses that could result in unfair treatment of individuals or groups (e.g., stereotyping, quality of service issues) or other undesirable harms (e.g., financial harms, legal risks) If so, please provide a description. Is there anything a future user could do to mitigate these undesirable harms?}

\textbf{A4:} No.

\noindent \textbf{5. Are there tasks for which the dataset should not be used? If so, please provide a description.}

\textbf{A5:} No.

\subsection{Distribution}

\noindent \textbf{1. Will the dataset be distributed to third parties outside of the entity (e.g., company, institution, organization) on behalf of which the dataset was created? If so, please provide a description.}

\textbf{A1:} Yes. The dataset will be made publicly available to the research community.

\noindent \textbf{2. How will the dataset will be distributed (e.g., tarball on website, API, GitHub)? Does the dataset have a digital object identifier (DOI)?}

\textbf{A2:} It will be publicly available on the project website at \href{https://github.com/JizhiziLi/RIM}{https://github.com/JizhiziLi/RIM}.

\noindent \textbf{3. When will the dataset be distributed?}

\textbf{A3:} The dataset will be distributed once the paper is accepted after peer-review.

\noindent \textbf{4. Will the dataset be distributed under a copyright or other intellectual property (IP) license, and/or under applicable terms of use (ToU)? If so, please describe this license and/or ToU, and provide a link or other access point to, or otherwise reproduce, any relevant licensing terms or ToU, as well as any fees associated with these restrictions.}

\textbf{A4:} It will be distributed under the CC BY-NC license.

\noindent \textbf{5. Have any third parties imposed IP-based or other restrictions on the data associated with the instances? If so, please describe these restrictions, and provide a link or other access point to, or otherwise reproduce, any relevant licensing terms, as well as any fees associated with these restrictions.}

\textbf{A5:} No.

\noindent \textbf{6. Do any export controls or other regulatory restrictions apply to the dataset or to individual instances? If so, please describe these restrictions, and provide a link or other access point to, or otherwise reproduce, any supporting documentation.}

\textbf{A6:} No.

\subsection{Maintenance}

\noindent \textbf{1. Who will be supporting/hosting/maintaining the dataset?}

\textbf{A1:} The authors.

\noindent \textbf{2. How can the owner/curator/manager of the dataset be contacted (e.g., email address)?}

\textbf{A2:} They can be contacted via email available on the project website.

\noindent \textbf{3. Is there an erratum? If so, please provide a link or other access point.}

\textbf{A3:} No. 

\noindent \textbf{4. Will the dataset be updated (e.g., to correct labeling errors, add new instances, delete instances)? If so, please describe how often, by whom, and how updates will be communicated to users (e.g., mailing list, GitHub)?}

\textbf{A4:} No.

\noindent \textbf{5. Will older versions of the dataset continue to be supported/hosted/maintained? If so, please describe how. If not, please describe how its obsolescence will be communicated to users.}

\textbf{A5:} N/A.

\noindent \textbf{6. If others want to extend/augment/build on/contribute to the dataset, is there a mechanism for them to do so? If so, please provide a description. Will these contributions be validated/verified? If so, please describe how. If not, why not? Is there a process for communicating/distributing these contributions to other users? If so, please provide a description.}

\textbf{A6:} N/A.